\documentclass[10pt,twocolumn,letterpaper]{article}

\usepackage[pagenumbers]{cvpr} 

\definecolor{cvprblue}{rgb}{0.21,0.49,0.74}
\usepackage[pagebackref,breaklinks,colorlinks,allcolors=cvprblue]{hyperref}
\usepackage{multirow}
\usepackage{algorithm}
\usepackage{algorithmicx}
\usepackage{algpseudocode}
\usepackage{enumitem}
\usepackage{pifont}
\usepackage{listings}

\def\paperID{} 
\def\confName{}
\def\confYear{}

\title{Real2Edit2Real: Generating Robotic Demonstrations via a 3D Control Interface}

\author{
Yujie Zhao$^{1,2\ast}$ \quad Hongwei Fan$^{1,2\ast}$ \quad Di Chen$^{3}$ \quad Shengcong Chen$^{3}$ \quad Liliang Chen$^{3}$ \quad Xiaoqi Li$^{1,2}$ \\ Guanghui Ren$^{3}$ \quad Hao Dong$^{1,2\dagger}$\\[1mm]
$^{1}$CFCS, School of Computer Science, Peking University\quad $^{2}$PKU-AgiBot Lab\quad $^{3}$AgiBot \\[1mm]
\textbf{\url{https://real2edit2real.github.io/}}
}
\begin{document}
\twocolumn[{
    \renewcommand\twocolumn[1][]{#1}
    \maketitle
    \vspace{-1.5em}
    \begin{center}
        \includegraphics[width=\textwidth]{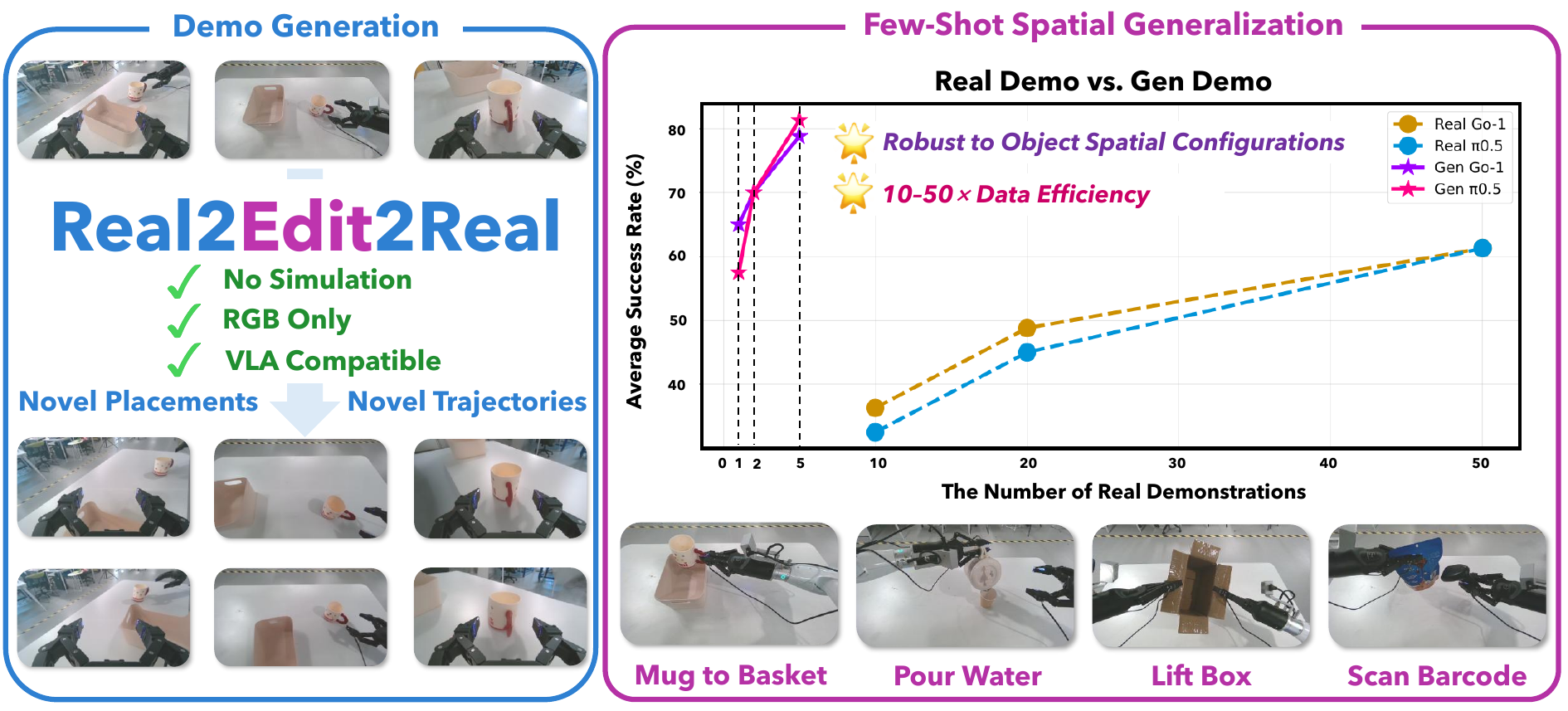}
        \vspace{-10pt}
        \captionof{figure}{
            The overview of Real2Edit2Real.
            Real2Edit2Real generates diverse robotic demonstrations, featuring 10-50$\times$ improvement on data efficiency compared with real-world collection across four tasks.
        }
        \label{fig:teaser}
    \end{center}
}]
\begin{abstract}
\renewcommand{\thefootnote}{}
\footnote{*: Equal contributions. $\dagger$: Corresponding author.}
Recent progress in robot learning has been driven by large-scale datasets and powerful visuomotor policy architectures, yet policy robustness remains limited by the substantial cost of collecting diverse demonstrations, particularly for spatial generalization in manipulation tasks. To reduce repetitive data collection, we present Real2Edit2Real, a framework that generates new demonstrations by bridging 3D editability with 2D visual data through a 3D control interface. Our approach first reconstructs scene geometry from multi-view RGB observations with a metric-scale 3D reconstruction model. Based on the reconstructed geometry, we perform depth-reliable 3D editing on point clouds to generate new manipulation trajectories while geometrically correcting the robot poses to recover physically consistent depth, which serves as a reliable condition for synthesizing new demonstrations. Finally, we propose a multi-conditional video generation model guided by depth as the primary control signal, together with action, edge, and ray maps, to synthesize spatially augmented multi-view manipulation videos. Experiments on four real-world manipulation tasks demonstrate that policies trained on data generated from only $1$–$5$ source demonstrations can match or outperform those trained on 50 real-world demonstrations, improving data efficiency by up to $10$-$50\times$. Moreover, experimental results on height and texture editing demonstrate the framework’s flexibility and extensibility, indicating its potential to serve as a unified data generation framework. Project website is \url{https://real2edit2real.github.io/}.
\end{abstract}    
\section{Introduction}
\label{sec:intro}

Recent advances in robotics have demonstrated remarkable progress in visuomotor policy learning, driven by large-scale datasets and powerful model architectures such as the Diffusion Policy~\cite{chi2023diffusionpolicy, chi2024diffusionpolicy} and Vision-Language-Action (VLA) models~\cite{rt22023arxiv, kim24openvla, black2024pi0, intelligence2025pi05, agibot2025agibotworld}. However, the performance of these methods heavily relies on the availability of diverse and high-quality demonstrations. 
Spatial generalization, in particular, remains a bottleneck for policy robustness~\cite{XueZ25demogen, saxena2025what}. In manipulation tasks where objects are randomly arranged in space, achieving a high success rate typically requires a large amount of data to cover diverse spatial configurations, and increases substantial data collection costs.

To alleviate the burden of repetitive data collection, an efficient strategy is to synthesize new demonstrations from limited existing data. MimicGen-style works~\cite{mandlekar2023mimicgen, garrett2024skillmimicgen, jiang2024dexmimicen, Hoque2024IntervenGen} segmented demonstration trajectories according to object interactions, and then transformed and interpolated these object-centric segments to generate new trajectories that adapt to novel object arrangements. Real2Render2Real~\cite{yu2025realrenderreal} synthesized demonstrations from a human manipulation video through pose tracking and trajectory interpolation. Since these approaches render robotic videos within a graphics engine, they inevitably face the visual and physics gaps, which remain a significant challenge in robotics. Moreover, they require assets for the manipulated objects, which prevents them from directly augmenting an existing demonstration. DemoGen~\cite{XueZ25demogen} augmented real-world point-cloud demonstrations through 3D editing and enhanced spatial generalization of 3D Diffusion Policy~\cite{Ze2024DP3}, but it cannot be applied to RGB images and 2D policies, which remain the dominant paradigm in current robot learning and deployment. Consequently, to our best knowledge, there is no existing method to rapidly scale up real-world 2D multi-view manipulation videos with novel trajectories, while preserving both visual realism and interaction fidelity.

To mitigate this research gap, we introduce \textbf{Real2Edit2Real}, a demonstration generation framework that bridges the gap between 3D editability and 2D visual data via a 3D control interface, achieving spatially augmented multi-view robotic demonstrations. 
As shown in Figure~\ref{fig:teaser}, Real2Edit2Real does not rely on simulation engines or digital assets, and directly generates multi-view manipulation data from raw RGB demonstrations, featuring novel object placements and corresponding new trajectories, which can be used for VLA training. 
Our key insight is that depth inherently encodes robot motion and object interactions, making it a natural interface between 3D modalities and 2D observations.
Specifically, our proposed framework works with three modules: 
(1) \textbf{Metric-scale geometry reconstruction} of robot manipulation scenarios, where we propose a hybrid training paradigm that leverages real and simulated data to co-train a feed-forward 3D reconstruction model. 
(2) \textbf{Depth-reliable spatial editing} which combines point-cloud editing with trajectory planning to generate feasible manipulation trajectories while geometrically correcting the robot’s poses, thereby producing kinematically consistent depth maps that serve as reliable control signals for subsequent video generation.
(3) \textbf{3D-Controlled video generation} for multi-view consistent demonstrations, where we construct a video generation model conditioned on depth, together with edges, actions, and ray maps. 
With Real2Edit2Real, we can edit 2D videos through a unified 3D control interface, which facilitates data augmentation for robotic manipulation, thereby enhancing the robustness of downstream policies.

To evaluate the quality and efficiency of generated demonstrations, we conduct experiments on four real-world robotic manipulation tasks, covering single-arm to dual-arm manipulation. Experimental results indicate that policies trained on data generated from as few as 1–5 source demonstrations can achieve comparable or higher success rates than those trained on 50 real-world demonstrations, improving data efficiency by up to 10–50$\times$. Additionally, Real2Edit2Real supports extended editing such as object height and background texture, demonstrating the framework’s flexibility and extensibility, and suggesting its potential as a unified robotic data generation framework.

\section{Related Work}
\label{sec:related_work}

\subsection{Demonstration Generation}

Due to the difficulty and cost of collecting robotic demonstration data, generating numerous demonstrations from zero or one demonstration has been proposed to rapidly extend robotic data. One line of works~\cite{agibot2025agibotworld,deng2025graspvla,villasevil2024reconciling,qureshi2025splatsim,han2025re} use simulation engines to automatically collect demonstrations with pre-defined tasks and motion planning. However, the lack of real-world interaction leads to the Sim2Real gap. Another line of works~\cite{mandlekar2023mimicgen,garrett2024skillmimicgen,jiang2024dexmimicen,Hoque2024IntervenGen,li2025momagen,XueZ25demogen,yu2025realrenderreal,robosplat} focus on generating from one or a few collected demonstrations. MimicGen family~\cite{mandlekar2023mimicgen,garrett2024skillmimicgen,jiang2024dexmimicen,Hoque2024IntervenGen,li2025momagen} generates from one demonstration with carefully designed task segments. DemoGen~\cite{XueZ25demogen} combines MimicGen-style generation with point cloud editing. However, it is incompatible with the widely used setting of multi-view RGB cameras for visuomotor policy training. Real2Render2Real~\cite{yu2025realrenderreal} and RoboSplat~\cite{robosplat} use 3D Gaussian Splatting (3DGS)~\cite{3dgs} with trajectory generation to reduce the gaps in visual fidelity and interaction reality. These works expose two disadvantages. First, the rendering-based techniques that they used still bring the visual domain gap, limiting the Gen2Real performance. Second, the dense image captures that 3DGS requires restrict the scalability of generation. By bridging point-cloud-based demonstration editing and 3D-controlled video generation, Real2Edit2Real jointly achieves scalability, visual quality, and real-world interaction of generated demonstrations in one framework.
\begin{figure*}[t]
    \centering
    \includegraphics[width=\textwidth]{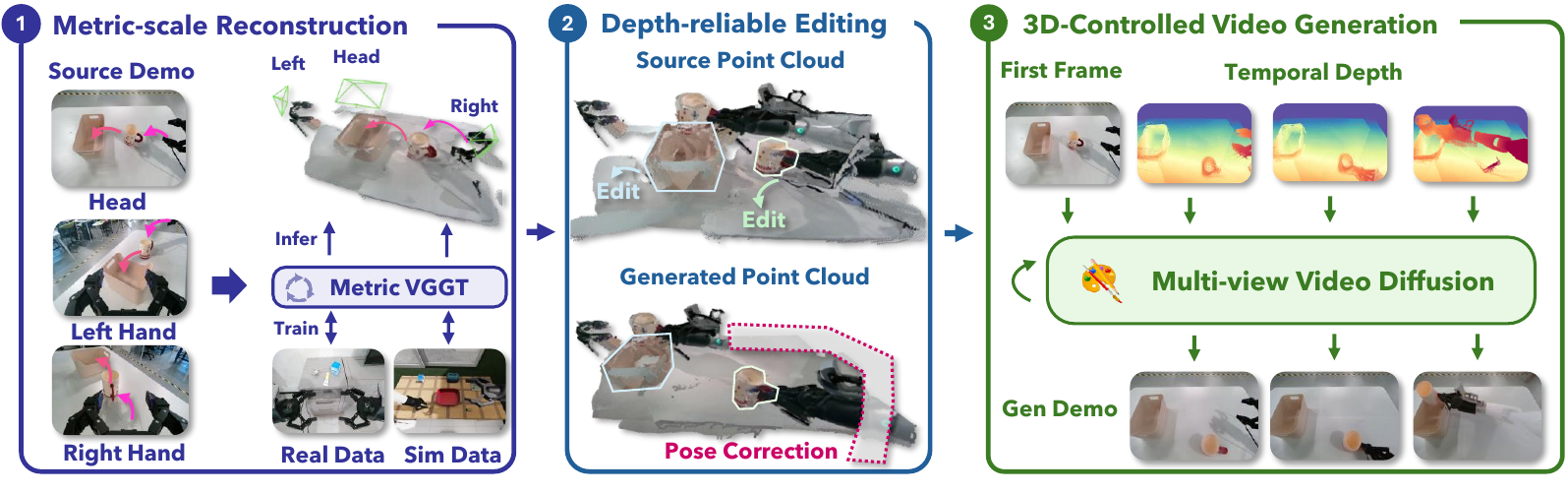}
    \caption{The framework of Real2Edit2Real. (1) Reconstruct metric-scale geometry from multi-view observations. (2)Synthesize novel trajectories with reliable depth rendering. (3) Generate demonstrations controlled by temporal depth signals.}
    \label{fig:method}
    \vspace{-15pt}
\end{figure*}

\subsection{Geometry Reconstruction}

Reconstructing the detailed environmental geometry is the key to generating controllable and multi-view consistent demonstrations. Early methods such as NeuS~\cite{wang2021neus} and 2DGS~\cite{huang20242dgs} use radiance fields and gaussian splatting as 3D representations, which require dense image captures and minute-level post-optimization, both restricting their application scope and efficiency. Recently, feed-forward geometry reconstruction~\cite{dust3r_cvpr24,mast3r_eccv24,wang2025vggt,wang2025pi3,keetha2025mapanything,lu2025matrix3d,zhang2025flare} unlocks sparse-view reconstruction in seconds, bringing the potential of recovering geometry from robot camera views~\cite{qian2025wristworld,wu2025geometry,lin2025evo}. However, directly using the feed-forward model suffers from the domain gap, including wrong camera pose and misaligned metrics between pretrained data and real-world manipulation scenarios. We propose an effective training recipe of VGGT~\cite{wang2025vggt} on robotic data to both utilize multi-view consistent depth maps and camera poses from simulated data, and precise geometry metrics from real-world data. The aligned geometry prediction ensures a high-quality point cloud, which in turn enables the generation of reliable depth maps for demonstration synthesis.

\subsection{Video Generation in Robotics}

Recent progress in video generation~\cite{videoworldsimulators2024, KlingAI, gao2025seedance, nvidia2025worldsimulationvideofoundation} has improved conditional and temporal consistency, empowering several downstream applications in robotics. First, predicting video generation can serve as an enhancement~\cite{tianpredictive, lu2024manigaussian, huang2025enerverse, zhao2024vlmpc} to regular policy learning by jointly predicting future action and observation in one model. Second, action-conditioned video generation~\cite{jiang2025enerverse, liao2025genie, wang2025embodiedreamer, shang2025roboscape} performs as the policy evaluator or realistic simulation environment, which receives future action and feeds generated observation back to the pretrained policy model. Third, learning multi-modal conditioned video generation~\cite{liu2025geometry, tong2025fidelity,liu2025robotransfer, qian2025wristworld,xu2025egodemogen} and editing these conditions during inference generalizes the original demonstration across diverse camera poses and textures. However, these works neglect spatial generalization, which is also fundamental for scaling robotic data. In contrast, Real2Edit2Real generates spatially augmented multi-view demonstrations and can be easily combined with first-frame editing to support beyond. 
\section{Method}
\label{sec:method}

\subsection{Overview}

We consider a humanoid robot scenario with multi-view videos from the head, left wrist, and right wrist cameras. Then we can formulate our problem as follows:
\begin{equation}
\mathcal{O} = \{\mathbf{O}_i\}_{i=1}^N = \text{Real2Edit2Real}(\mathbf{O}, \mathcal{K})
\label{eq:problem}
\end{equation}

Here, $\mathbf{O} = (\mathbf{I}, \mathbf{q}, \mathbf{a})$ consists of multiview videos $\mathbf{I} = \{(I_h^t, I_l^t, I_r^t)\}_{t=1}^{N}$, joint angles $\mathbf{q}$, and actions $\mathbf{a}$ from one source demonstration. $\mathcal{K}$ represents the robot kinematic model (URDF) and camera parameters, where $\mathcal{K} = (\mathcal{K}_{robot}, \mathcal{K}_{cam})$. With our proposed Real2EditReal framework, the source demonstration $\mathbf{O}$ can be augmented to a large set of demonstrations $\mathcal{O}$ with novel object spatial configurations and trajectories.

As shown in Figure~\ref{fig:method}, our framework consists of three main components: metric-scale geometry reconstruction, depth-reliable spatial editing, and 3D-controlled video generation. In Sec.~\ref{subsec:reconstruct}, we first introduce a hybrid training paradigm that leverages both real and simulated data to enhance the capability of the reconstruction model $\mathcal{R}$ in robotic environments, which predicts depth maps $\mathbf{D}$ and camera poses $\mathbf{T}$ from the source demo as Eq.~\ref{eq:recon_def}, where $\mathbf{D} = \{(D_h^t, D_l^t, D_r^t)\}_{t=1}^{N}$ and $\mathbf{T} = \{(T_h^t, T_l^t, T_r^t)\}_{t=1}^{M}$.
\begin{equation}
    \mathbf{D}, \mathbf{T} = \mathcal{R}(\mathbf{I})
\label{eq:recon_def}
\end{equation}
In Sec.~\ref{subsec:edit}, we detail the pipeline $\mathcal{E}$ of depth-reliable spatial editing, shown in Eq.~\ref{eq:edit_def}, where we synthesize novel trajectories based on motion planning and point-cloud editing, while correcting the robot’s poses to obtain physically consistent depth maps that serve as reliable control signals.
\begin{equation}
    \{(\mathbf{D}_{i}, \mathbf{T}_{i}, \mathbf{q}_{i}, \mathbf{a}_{i})\}_{i=1}^{N} = \mathcal{E}(\mathbf{D}, \mathbf{T}, \mathcal{K}, \mathbf{q}, \mathbf{a})
\label{eq:edit_def}
\end{equation}
In Sec.~\ref{subsec:generate}, we propose a multi-view video generation model $\mathcal{G}$, which produces the complete robotic manipulation video from the first frame controlled by depth, together with Canny edges $\mathcal{C}(\cdot)$, actions, and ray maps, as Eq.~\ref{eq:gen_def}.
\begin{equation}
    \{\mathbf{I}_{i}\}_{i=1}^{N} = \{\mathcal{G}(\mathbf{D}_{i}, \mathbf{T}_{i}, \mathbf{a}_{i}, \mathcal{C}(\mathbf{D}_i))\}_{i=1}^{N}
\label{eq:gen_def}
\end{equation}

\subsection{Metric-scale Geometry Reconstruction}
\label{subsec:reconstruct}
Motivated by the fact that 3D data affords greater flexibility for editing than 2D imagery, we initially conduct geometric reconstruction of the source demonstrations.
To achieve scale-aware geometry reconstruction and improve its accuracy in humanoid robot scenes, which only include three cameras from the head and both wrists, we propose a hybrid training paradigm that combines real and simulated data to fine-tune VGGT~\cite{wang2025vggt}, enhancing metric-scale depth map and camera pose prediction in humanoid scenarios. 

\noindent\textbf{Camera Pose.} Camera poses obtained via hand–eye calibration in real-robot settings are susceptible to mechanical tolerances, calibration drift, and kinematic inaccuracies, leading to misalignment. Conversely, simulated data offers perfectly accurate and metrically consistent camera poses, since the virtual sensors are derived directly from the robot’s standardized URDF model. Therefore, we only use simulated data to supervise the camera loss: $\mathcal{L}_{\text{camera}} = \sum_{v \in \{h, l, r\}}
\mathcal{L}_1(\hat{T}_v^{\text{sim}}, T_v^{\text{sim}})$, where we use $\mathcal{L}_1$ loss because of the precise ground truth. 

\noindent\textbf{Depth Map.} Real-world datasets capture metric-scale geometry through depth sensors, but the acquired depth maps are often highly noisy due to sensor limitations, reflective or textureless surfaces. In contrast, simulated data provides noise-free and geometrically precise depth maps, but the object and scene scale may deviate from real-world distributions. To leverage the complementary strengths of both fields of data, we compute the depth loss using a mixed method:
$\mathcal{L}_{\text{depth}} = \sum_{v \in \{h, l, r\}} ((\mathcal{L}_\text{conf}(\mathcal{M}(\hat{D}_v^{real}), \mathcal{M}(D_v^{real}))+\mathcal{L}_\text{conf}(\hat{D}_v^{sim}, D_v^{sim}))$. 
$\mathcal{L}_\text{conf}$ is the depth loss with uncertainty used in~\cite{wang2025vggt}, and we use a threshold mask $\mathcal{M}$ to filter invalid noise in real depth maps.

In addition, the simulation data also supervises the point map loss:  
$\mathcal{L}_{\text{pointmap}} = \sum_{v \in \{h, l, r\}} \mathcal{L}_\text{conf}(\hat{P}_v^{sim}, P_v^{sim})$. 
The total training loss is shown as Eq.\ref{eq:loss}, where we apply the weight $\lambda$ of 10 to $\mathcal{L}_\text{camera}$ to ensure that its magnitude is comparable to other losses, which stabilizes optimization.
\begin{equation}
    \mathcal{L} =\lambda\mathcal{L}_\text{camera} + \mathcal{L}_\text{depth} + \mathcal{L}_\text{pointmap}
\label{eq:loss}
\end{equation}

\subsection{Depth-reliable Spatial Editing}
\label{subsec:edit}
Based on point-cloud editing and motion planning, we can synthesize novel object placements and corresponding manipulation trajectories in 3D space. To obtain reliable depth from the edited point clouds, we improve the spatial editing pipeline with techniques such as background inpainting and depth filtering. Crucially, we introduce robot pose correction during spatial editing, ensuring the resulting depth maps are consistent with the robot’s kinematics.

\begin{algorithm}[thbp]
\caption{Depth-reliable Spatial Editing Pipeline}
\label{alg:edit}
\begin{algorithmic}
\Require Source point clouds $\mathcal{P}^{robot}, \mathcal{P}^{object},\mathcal{P}^{bg}$, joint states $\mathcal{Q}$, action trajectory $\mathcal{A}$, camera poses $\mathcal{T}$. 
\Ensure Novel depth sequence $\mathcal{D}^\star$, joint states $\mathcal{Q}^\star$, action trajectory $\mathcal{A}^\star$, camera poses $\mathcal{T}^\star$. 

\Function{RenderDepth}{$\mathcal{P}, \mathcal{T}, \mathcal{Q}$}
    \State $D_1 \gets \text{ProjectPointCloud}(\mathcal{P}, \mathcal{T})$
    \State $D_2 \gets \text{RenderLinkDepth}(\mathcal{Q}, \mathcal{T})$
    \State \Return $\text{Merge}(D_1, D_2)$
\EndFunction
\State Random Sample a Object Transform $\mathbf{T} \in \mathbb{R}^{4\times 4}$
\State $\mathcal{D}^\star \gets list()$,
$\mathcal{Q}^\star \gets list()$, 
$\mathcal{A}^\star \gets list()$, 
$\mathcal{T}^\star \gets list()$ 
\State // Motion Segment
\State $\mathcal{A}^\star_{start} \gets \mathcal{A}_0$, 
$\mathcal{A}^\star_{end} \gets \mathbf{T}A_{end}$,
\For{$t$ in Motion}
    \State $\mathbf{T}_t, \mathcal{A}^\star_t,\mathcal{Q}^\star_t \gets \text{MotionPlan}(\mathcal{A}^\star_{start}, \mathcal{A}^\star_{end}, t)$
    \State $\mathcal{P}^{arm}_t \gets \text{FK}(\mathcal{P}^{robot}_t, \mathcal{Q}_t)$
    \State $\mathcal{P}^{ee}_t \gets \mathcal{P}^{robot}_t \setminus \mathcal{P}^{arm}_t$
    \State $\mathcal{P}^\star_t \gets \mathbf{T}_t\mathcal{P}^{ee}_t \cup \mathbf{T}\mathcal{P}^{object}_t \cup \mathcal{P}^{bg}$
    \State $\mathcal{D}^\star \gets \mathcal{D}^\star \cup \text{RenderDepth}(\mathcal{P}^\star_t, \mathbf{T}_t\mathcal{T}_t, \mathcal{Q}^\star_t)$
    \State $\mathcal{Q^\star} \gets  \mathcal{Q^\star} \cup \mathcal{Q}^\star_t$, $\mathcal{A}^\star \gets \mathcal{A}^\star \cup \mathcal{A}^\star_t$, 
    $\mathcal{T}^\star \gets \mathcal{T}^\star \cup \mathbf{T}_t\mathcal{T}_t$
\EndFor
\State // Skill Segment
\For{$t$ in Skill}
    \State $\mathcal{Q}^\star_t \gets \text{IK}(\mathbf{T}\mathcal{A}_t)$
    \State $\mathcal{P}^{arm}_t \gets \text{FK}(\mathcal{P}^{robot}_t, \mathcal{Q}_t)$
    \State $\mathcal{P}^{ee}_t \gets \mathcal{P}^{robot}_t \setminus \mathcal{P}^{arm}_t$
    \State $\mathcal{P}^\star_t \gets \mathbf{T}(\mathcal{P}^{ee}_t \cup \mathcal{P}^{object}_t) \cup \mathcal{P}^{bg}$
    \State $\mathcal{D}^\star \gets \mathcal{D}^\star \cup \text{RenderDepth}(\mathcal{P}^\star_t, \mathbf{T}\mathcal{T}_t, \mathcal{Q}^\star_t)$
    \State $\mathcal{Q^\star} \gets  \mathcal{Q^\star} \cup \mathcal{Q}^\star_t$, $\mathcal{A}^\star \gets \mathcal{A}^\star \cup \mathbf{T}\mathcal{A}_t$, 
    $\mathcal{T}^\star \gets \mathcal{T}^\star \cup \mathbf{T}\mathcal{T}_t$
\EndFor

\State \Return $\mathcal{D}^\star$, $\mathcal{Q}^\star$, $\mathcal{A}^\star$, $\mathcal{T}^\star$ 
\end{algorithmic}
\end{algorithm}

\begin{figure*}[t]
    \centering
    \includegraphics[width=\textwidth]{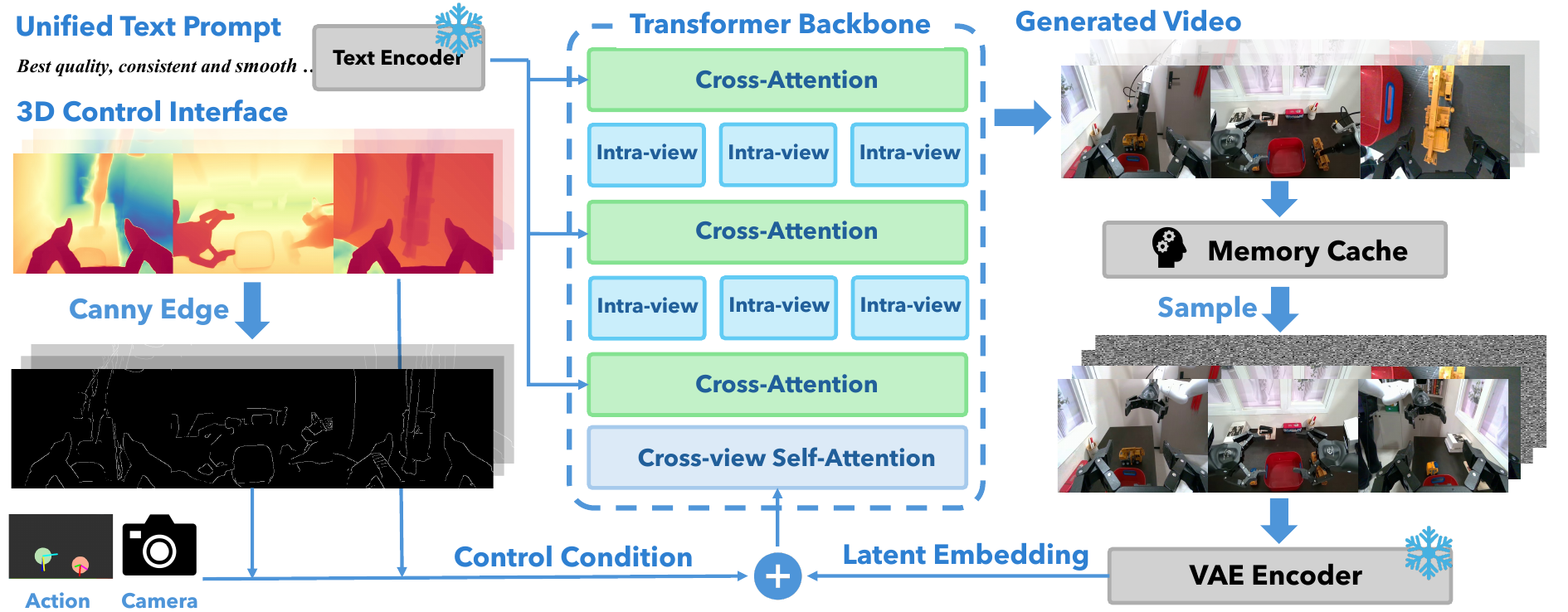}
    \caption{The framework of 3D-Controlled Video Generation. We utilize depth as the 3D control interface, in conjunction with edges, actions, and ray maps, to guide the generation of multi-view demonstrations.}
    \label{fig:model}
    \vspace{-15pt}
\end{figure*}

\noindent\textbf{Trajectory Synthesis.} Inspired by the previous work~\cite{XueZ25demogen}, we decompose a point-cloud demonstration into two types of segments: the motion segment, where the robot moves freely in space, and the skill segment, where the robot interacts with objects. Given a transformation that relocates an object, we apply the same transformation to the robot’s point cloud in the corresponding skill segment. This ensures that the robot-object relation remains consistent with the original demonstration, thereby preserving realistic interactions. The new motion segment is generated through motion planning~\cite{sundaralingam2023curobo}.

\noindent\textbf{Depth Projection.} Since the camera is rigidly attached to the robot’s end-effector, we apply the same transformation to the camera pose to obtain its updated position after editing. This allows us to project the edited point cloud to generate the corresponding depth map. However, the projected depth maps often suffer from holes and noise due to changes in object positions and varying camera distances, which can cause sparse or missing regions. We mitigate these artifacts through background~\cite{wang2025seededit30fasthighquality} inpainting and depth filtering.

\noindent\textbf{Robot Pose Correction.} A major challenge arises from incorrect robot poses, as the previous editing process treats the entire robot as a rigid body. Instead, only the end-effector should be transformed, and the remaining arm must be realigned to preserve kinematic validity. To ensure the reliability of depth maps, we segment out the original robot links with URDF and source joint states, and re-render the arm depth with synthesized actions. This process yields physically plausible robot configurations and produces accurate depth observations without rigid-body artifacts. To summarize the overall procedure, we illustrate in Algorithm~\ref{alg:edit} an example pipeline consisting of one motion segment followed by one skill segment.

\subsection{3D-Controlled Video Generation}
\label{subsec:generate}
After obtaining the edited depth, action, and camera pose sequences from the 3D editing pipeline described in Section~\ref{subsec:edit}, we then convert them into 2D visual observations required for policy training. We propose a 3D-conditioned video generation model that, starting from the first frame, synthesizes novel robot manipulation videos with realistic visual appearance, multi-view consistency, and physically plausible interactions. As shown in Figure~\ref{fig:model}, we build a Transformer-based video generation model followed by~\cite{peebles2023@dit, nvidia2025worldsimulationvideofoundation, liao2025genie}. The model works with three key designs: dual-attention mechanism, depth control interface, and smooth object relocation.

\noindent\textbf{Dual-attention Mechanism.} The dual-attention mechanism consists of \textit{intra-view attention} and \textit{cross-view attention}. Intra-view attention performs self-attention over the tokens of each individual view, capturing detailed spatial context within that view. Cross-view attention, on the other hand, computes self-attention across all views simultaneously, enabling the model to utilize the multi-view correspondence. This dual-attention design not only facilitates multi-view consistency in the generated videos by allowing interactions of visual features across different scale contexts, but also significantly reduces computational cost compared to applying global attention at every layer. 

\noindent\textbf{Depth Control Interface.} We use depth as the 3D control interface of video generation. Specifically, we concatenate the depth map with the image latent representations and feed them jointly into the transformer backbone, enabling the model to condition video generation on 3D structural cues. This design ensures the synthesized robotic demonstrations remain consistent with the geometry information in the depth sequence, which encodes robot motion and object interactions.
In addition, we incorporate auxiliary conditioning signals, including Canny edges, action maps, and ray maps, which further sharpen object boundaries, improve motion grounding, and enhance multi-view consistency, respectively. Overall, these structured controlling signals provide strong 3D inductive biases, enabling more realistic manipulation behaviors in the generated videos.

\begin{table*}[t]
\renewcommand\tabcolsep{7pt}
\centering
\begin{tabular}{l|cccccccc|cc}
\toprule
\multirow{2}{*}{\textbf{\# Demo}} & \multicolumn{2}{c}{\textbf{Mug to Basket}} & \multicolumn{2}{c}{\textbf{Pour Water}} & \multicolumn{2}{c}{\textbf{Lift Box}} & \multicolumn{2}{c}{\textbf{Scan Barcode}} & \multicolumn{2}{|c}{\textbf{Total}}\\
\cmidrule(lr){2-11}
& Go-1 & $\pi_{0.5}$ & Go-1 & $\pi_{0.5}$ & Go-1 & $\pi_{0.5}$ & Go-1 & $\pi_{0.5}$ &  Go-1 & $\pi_{0.5}$ \\
\midrule
Real 10 & 8 / 20 & 8 / 20 & 5 / 20 & 1 / 20 & 11 / 20 & 13 / 20 & 5 / 20 & 4 / 20 & 36.3\% & 32.5\%\\
Real 20 & 12 / 20 & 14 / 20 & 7 / 20 & 2 / 20 & 12 / 20 & 15 / 20 & 8 / 20 & 5 / 20 & 48.8\% & 45.0\%\\
Real 50 & 14 / 20 & 13 / 20 & 8 / 20 & 8 / 20 & 15 / 20 & 17 / 20 & 12 / 20 & 11 / 20 & 61.3\% & 61.3\%\\
\midrule
Real 1 Gen 200 & 14 / 20 & 15 / 20 & 12 / 20 & 10 / 20 & 12 / 20 & 10 / 20 & 14 / 20 & 11 / 20 & 65.0\% & 57.5\% \\
Real 2 Gen 200 & 15 / 20 & 15 / 20 & 10 / 20 & 10 / 20 & 14 / 20 & 17 / 20 & 17 / 20 & 14 / 20 & 70.0\% & 70.0\% \\
Real 5 Gen 200 & 17 / 20 & 18 / 20 & 12 / 20 & 12 / 20 & 16 / 20 & 18 / 20 & 18 / 20 & 17 / 20 & \textbf{78.8\%} & \textbf{81.3\%} \\
\bottomrule
\end{tabular}
\caption{Success rates of Real2Edit2Real on four real-world manipulation tasks and two VLA policies Go-1 and $\pi_{0.5}$. Both policies trained on data generated from only 1–5 source demonstrations can match or outperform those trained on 50 real-world demonstrations, improving data efficiency by up to 10-50$\times$.}
\label{tab:real_robot_res}
\vspace{-15pt}
\end{table*}

\noindent\textbf{Smooth Object Relocation.} With the condition's control, the model is able to generate the manipulation video from the first frame, but how to relocate the objects in the first frame remains difficult. To make full use of the depth-controlled video generation model, we convert the object relocation to a smooth transformation, where we interpolate both the translation and rotation of objects during spatial editing to synthesize object moving trajectories before manipulation starts. Through this smooth relocation, we convert image editing to video generation and process the object relocation and demonstration generation together by 3D-controlled video generation, achieving a unified and efficient generation framework.

\section{Experiments}
\label{sec:experiment}

To demonstrate the effectiveness of our data generation framework, we present the following experiments. 
In Section~\ref{subsec:policy_result}, we evaluate its impact on real-world policy learning across four manipulation tasks. 
In Section~\ref{subsec:more_edit}, we showcase the flexibility of our proposed framework through height and texture editing. 
In Section~\ref{subsec:ablation}, we conduct ablations to validate the necessity of each module and our key designs.
Finally, in Section~\ref{subsec:vis}, we provide visualizations of our novel demonstration generation.

\subsection{Implementation Details}
\label{subsec:exp_detail}

\noindent\textbf{Real2Edit2Real.} For Metric-VGGT, we sample 40K frames from the Agibot-DigitalWorld dataset~\cite{contributors2025agibotdigitalworld} as simulation training data and collect 100K real robot data with depth sensors as real training data. We full-finetune VGGT~\cite{wang2025vggt} on 8 H100 GPUs for 150K iterations, 20 hours with the learning rate 2e-4 and backbone learning rate 2e-5. For the video generation model, we sample 7K episodes from 64 tasks in the AgiBot-World dataset~\cite{agibot2025agibotworld} as training data. We train the video generation model by fine-tuning the backbone of GE-Sim~\cite{liao2025genie}(based on Cosmos-Predict-2B~\cite{contributors2025agibotdigitalworld}) on 8 H100 GPUs for 20K iterations, 60 hours with the learning rate 1e-4, batch size 16. With parallelization across 8 H100 GPUs, the average generation time for a 20-second, 30-FPS episode is 48.6 seconds. Please refer to the supplementary materials for more details.
 
\noindent\textbf{Hardware Setup.} We use the Agibot Genie G1 robot with its internal motion API. Three RGB cameras are mounted on its head, left wrist, and right wrist. The workspace is a 50cm $\times$ 40cm rectangular area on the white desktop. Please refer to the supplementary materials for more details.

\noindent\textbf{VLA Policy.} We conduct experiments on two VLA policies: Go-1~\cite{agibot2025agibotworld} and $\pi_{0.5}$~\cite{intelligence2025pi05}. For Go-1, we only finetune the action expert while keeping the backbone frozen because we use the same embodiment as its pretrained data.  The action is the 6D end-effector pose. For $\pi_{0.5}$, due to embodiment mismatch, we perform full finetuning. The action is the 7-DoF joint angles. Each training typically consists of 10K iterations; for the smaller training data cases, we proportionally reduce the iteration count to 100 epochs. Both policies are trained on 8 H100 GPUs for 2-4 hours.

\subsection{Gen2Real Policy Learning}
\label{subsec:policy_result}

\textbf{Tasks.}
As shown in Figure~\ref{fig:teaser}, we conduct real-robot experiments on four tasks, covering from single-arm to dual-arm manipulation: 
\begin{itemize}
    \item Mug to Basket: The robot uses its right arm to grasp the mug and place it stably inside the basket.
    \item Pour Water: The robot uses its left arm to pick up the kettle and pour water into the paper cup by aligning the spout with the cup.
    \item Lift Box: The robot grasps both sides of the box using its two arms and lifts it.
    \item Scan Barcode: The robot grasps a snack with its left hand and a barcode scanner with its right hand, and scans the barcode by aligning the scanner with it.
\end{itemize}

\noindent\textbf{Settings.} For real-world data, we collect demonstrations via teleoperation by placing objects in diverse configurations that uniformly cover the workspace, including variations in both position and orientation.
For generation experiments, we randomly sample a specified number of source demonstrations from the collected data and apply randomized object relocations around their original poses. This procedure yields 200 synthesized demonstrations, and training is performed solely on these generated samples.
During evaluation, objects are uniformly randomly placed across the workspace to assess spatial generalization.

\noindent\textbf{Results.} Table~\ref{tab:real_robot_res} shows the manipulation success rate of four tasks with different training data. Real data results show that when the number of demonstrations is fewer than 20, the average task success rate drops below 50\%, indicating that VLAs exhibit limited spatial generalization when trained with scarce data. Conversely, policies trained on 200 demonstrations generated from only a single source demonstration by Real2Edit2Real achieve comparable spatial generalization to those trained on 50 real-world demonstrations. As the number of source demonstrations increases, the average success rate of policies trained on the same 200 generated demonstrations improves significantly. When trained with data generated from 5 real demonstrations, the policies of Go-1 and $\pi_{0.5}$ achieve average success rates of 78.8\% and 81.3\%, surpassing those trained on 50 real demonstrations by 17.5\% and 20\%, respectively. This improvement arises because increasing source demonstrations introduces more diverse robot–object interaction patterns and expands the spatial coverage of the generated data. Overall, the experimental results demonstrate that Real2Edit2Real improves data efficiency by $10$-$50\times$ through data generation, confirming its effectiveness as a demonstration generation framework for robotics.

\subsection{More Applications}
\label{subsec:more_edit}
\begin{figure}[t]
    \centering
    \includegraphics[width=\linewidth]{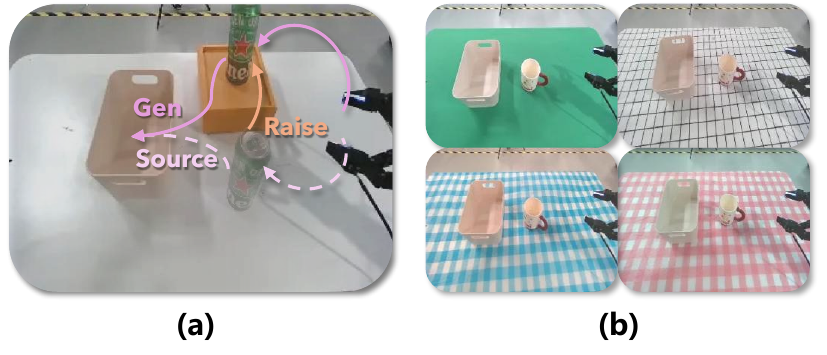}
    \caption{Experiment setups of (a) height and (b) texture editing.}
    \label{fig:height}
    \vspace{-10pt}
\end{figure}

\begin{table}[t]
\renewcommand\tabcolsep{6pt}
\centering
\begin{tabular}{lccc}
\toprule
\textbf{\# Demo} & \textbf{Tabletop} & \textbf{Platform} & \textbf{Total} \\
\midrule
Tabletop Real 20 & 5/5 & 0/5 & 50\% \\
Tabletop Real 1 Gen 40 &  4/5 & 4/5 & \textbf{80\%} \\
\bottomrule
\end{tabular}
\caption{Performance comparison of Go-1 on height generalization. Our method successfully generalizes to the unseen height.}
\label{tab:height}
\vspace{-10pt}
\end{table}

\noindent\textbf{Height Editing.} 
With smooth object relocation described in Sec.~\ref{subsec:generate}, we can also edit the object height as shown in Figure~\ref{fig:height}. Table~\ref{tab:height} shows the performance comparison on height generalization. The policy trained by 20 real demonstrations on the tabletop completely fails on the platform height because of OOD. If we generate 20 demonstrations on the tabletop and 20 demonstrations on the platform through our framework, the policy can achieve $80\%$ success rate.

\begin{table}[t]
\renewcommand\tabcolsep{2pt}
\centering
\begin{tabular}{lcccccc}
\toprule
\textbf{\# Demo} & \textbf{White} & \textbf{Green} & \textbf{Black} & \textbf{Blue} & \textbf{Red} & \textbf{Total}\\
\midrule
Real 50 & 7/10 & 4/10 & 5/10 & 5/10 & 4/10 & 50\% \\
Real 1 Gen 200 & 7/10 & 4/10 & 6/10 & 4/10 & 5/10 & 52\%\\
Real 1 Gen 200* & 6/10 & 7/10 & 6/10 & 7/10 & 8/10 & \textbf{68\%}\\
\bottomrule
\end{tabular}
\caption{Performance comparison of Go-1 under different desktop textures. Real 1 Gen 200* means generating data includes different textures. Our method is robust to the texture variation.}
\label{tab:texture}
\vspace{-10pt}
\end{table}

\begin{figure}[t]
    \centering
    \includegraphics[width=\linewidth]{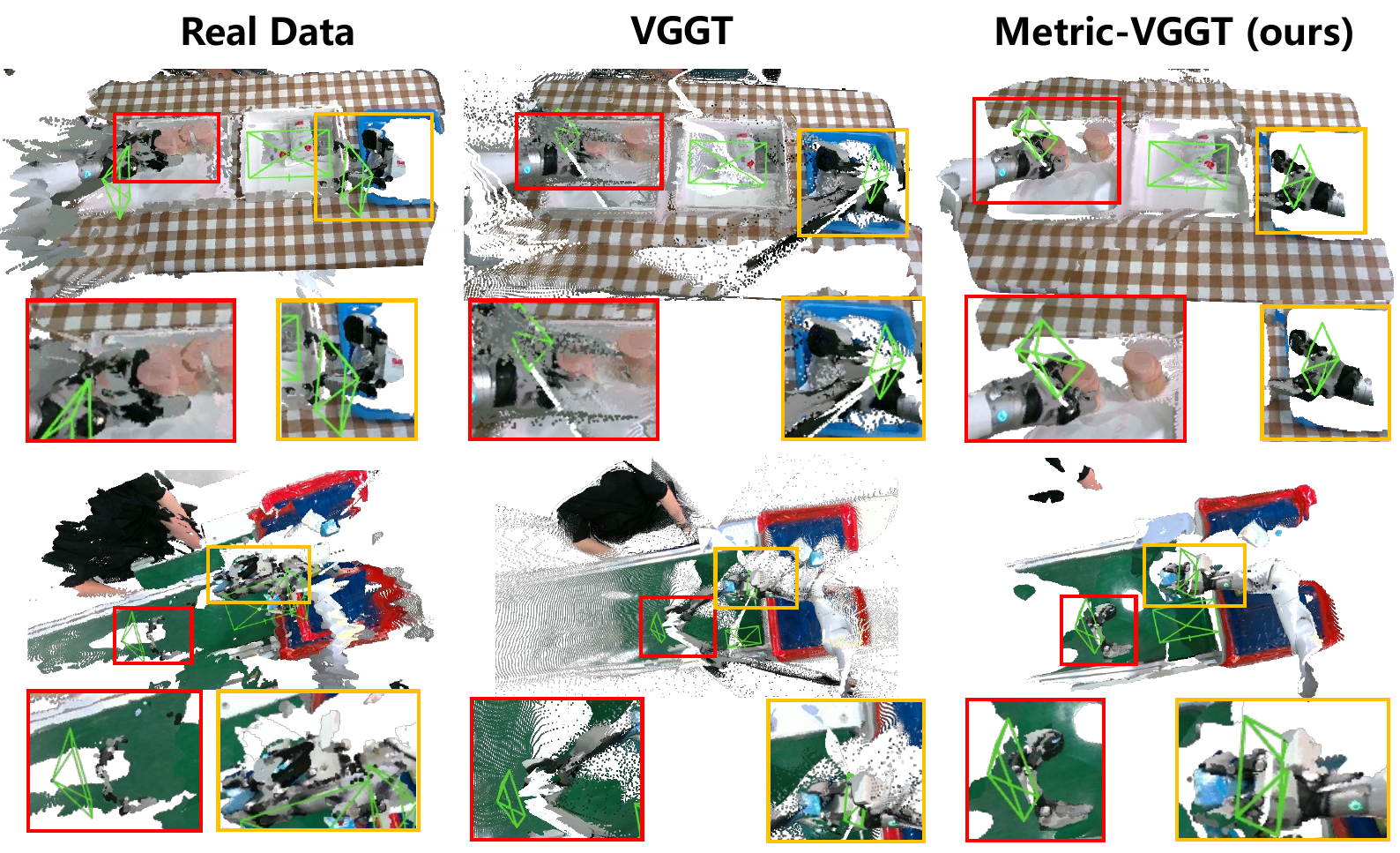}
    \caption{Ablation study of geometry reconstruction. The left end-effector is in \textcolor{red}{red}, and the right end-effector is in \textcolor{Goldenrod}{yellow}.}
    \vspace{-10pt}
    \label{fig:vggt_vis}
\end{figure}

\begin{figure}[t]
    \centering
    \includegraphics[width=\linewidth]{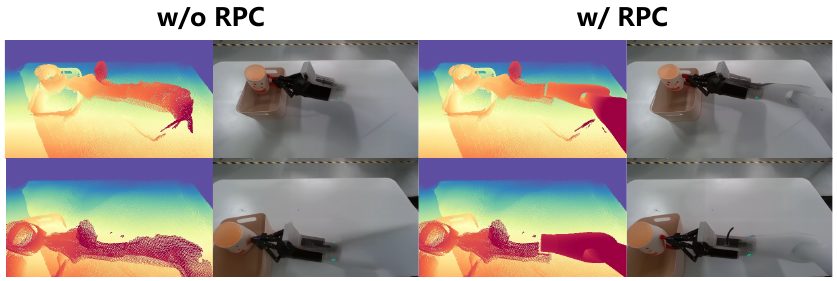}
    \caption{Ablation study of robot pose correction (RPC).}
    \label{fig:arm_render}
    \vspace{-10pt}
\end{figure}

\begin{figure}[t]
    \centering
    \includegraphics[width=\linewidth]{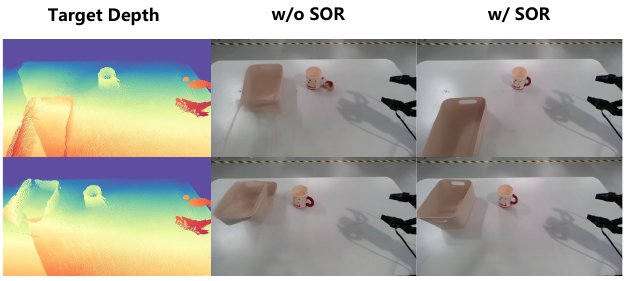}
    \caption{Ablation study of smooth object relocation (SOR).}
    \label{fig:sor}
    \vspace{-10pt}
\end{figure}

\begin{figure*}[t]
    \centering
    \includegraphics[width=0.9\textwidth]{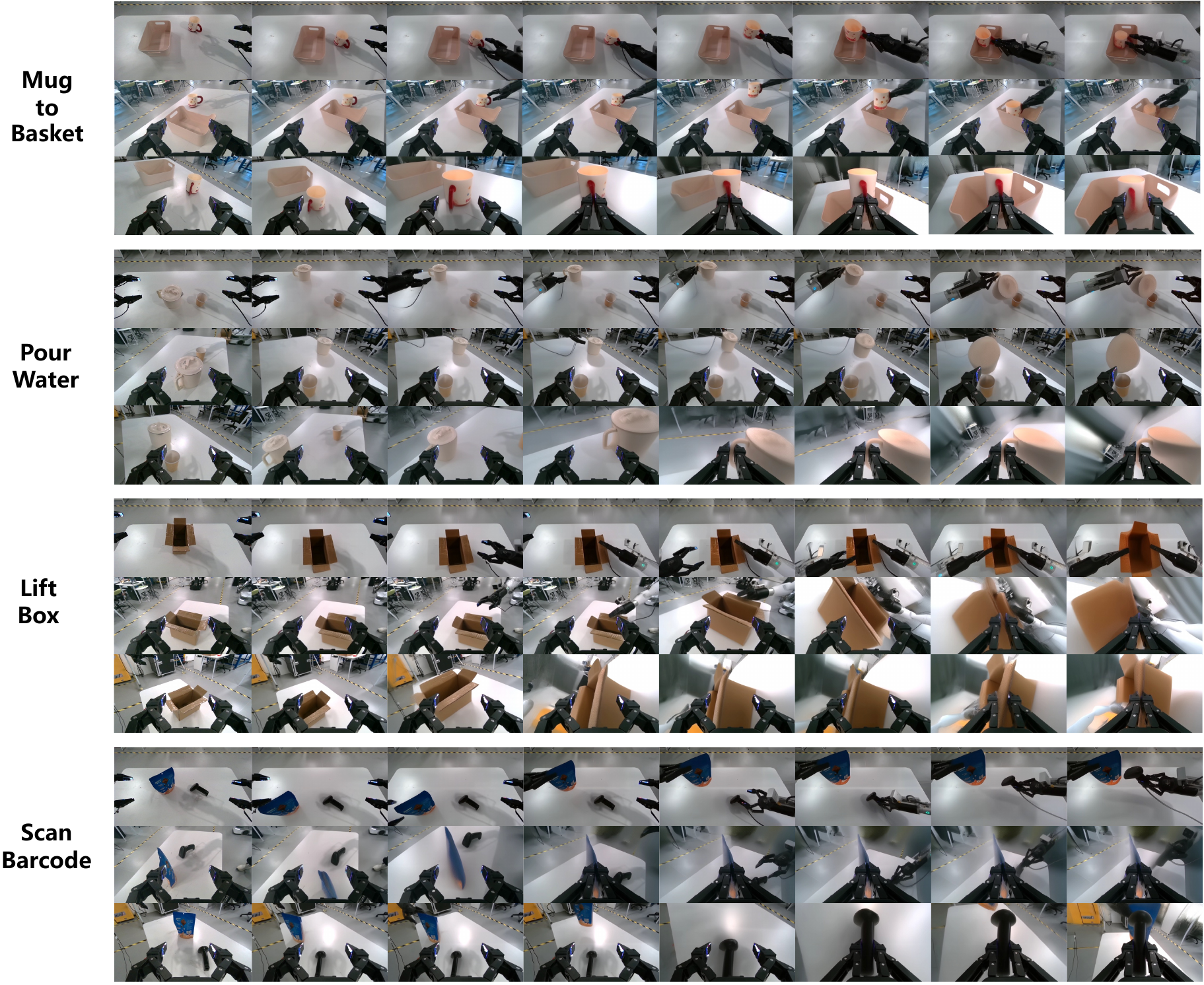}
    \caption{Visualization of videos generated by Real2Edit2Real on all four real-world tasks.}
    \label{fig:vis}
    \vspace{-15pt}
\end{figure*}

\noindent\textbf{Texture Editing.} 
Since the model described in Sec.~\ref{subsec:generate} generates video from the first frame, we can easily edit the video through first-frame editing, like changing the background texture. Table~\ref{tab:texture} illustrates the performance under different desktop textures shown in Figure~\ref{fig:height} and indicates that our method can generate demonstrations with different textures to improve policy robustness.

\subsection{Ablation Study}
\label{subsec:ablation}

\noindent\textbf{Geometry Reconstruction.} 
Figure~\ref{fig:vggt_vis} provides point-cloud visualizations of real data, VGGT, and ours. We can see that our model predicts the cleanest point cloud and the most accurate camera pose in robot scenarios, while the real-world data suffer from inaccurate camera poses and the VGGT reconstructions contain substantial clutter and noise.

\noindent\textbf{Robot Pose Correction.}
Figure~\ref{fig:arm_render} shows the ablation results of robot pose correction. Without robot pose correction, the erroneous depth maps lead to blurry and inconsistent generation results. With robot pose correction, the depth maps become kinematically consistent, allowing the model to generate realistic robot motions in the synthesized videos.

\noindent\textbf{Smooth Object Relocation.}
Figure~\ref{fig:sor} shows the ablation results of smooth object relocation. Without smooth object relocation, the generated object placements often exhibit noticeable errors, leading to unusable demonstrations. In contrast, smooth object relocation enables the precise placement of objects at the target locations.

\subsection{Visualization}
\label{subsec:vis}
Figure~\ref{fig:vis} is the visualization of videos generated by Real2Edit2Real on all four real-world tasks. The generated videos successfully relocate the objects, synthesize the correct manipulation trajectories, and maintain both multi-view consistency and realistic visual appearance. Please refer to the supplementary materials for more visualizations.

\section{Conclusion}
\label{sec:conclusion}

In this work, we introduce Real2Edit2Real, a framework that enables scalable demonstration generation by linking 3D editability with 2D visual data. Through metric-scale geometry reconstruction, depth-reliable spatial editing, and 3D-controlled video generation, our approach synthesizes realistic and kinematically consistent multi-view manipulation demonstrations. 
Experiments across four real-world manipulation tasks show that policies trained on data generated from as few as 1–5 demonstrations can match or surpass those trained on 50 real-world demonstrations, improving data efficiency by up to 10–50$\times$. Additional results on height and texture editing further highlight the extensibility of our framework, suggesting its potential to serve as a unified engine for scalable data generation.
\section{Acknowledgement}
\label{sec:ack}
\begin{sloppypar}
This work was supported by the National Natural Science Foundation of China (62376006). We would like to thank Zizhao Tong from University of Chinese Academy of Sciences for his fruitful discussion and Haolin Chen from Zhongguancun Academy for his technical support.
\end{sloppypar}
{
    \small
    \bibliographystyle{ieeenat_fullname}
    \bibliography{main}
}

\clearpage

\twocolumn[{
    \renewcommand\twocolumn[1][]{#1}
    \maketitlesupplementary
    \begin{center}
        \normalsize 
        \renewcommand\tabcolsep{6pt}
        \begin{tabular}{l|cc|ccc}
        \toprule
        \multirow{2}{*}{\textbf{Method}} & \multicolumn{2}{c|}{\textbf{Source}} &  \multicolumn{3}{c}{\textbf{Generation}} \\
        & \textbf{No Simulation} & \textbf{RGB Only} & \textbf{VLA Compatible} & \textbf{Novel Texture} & \textbf{Novel Trajectory}\\
        \midrule
        MimicGen~\cite{mandlekar2023mimicgen} & \ding{55} & \ding{55} & \ding{51} & \ding{51} & \ding{51} \\
        SkillMimicGen~\cite{garrett2024skillmimicgen} & \ding{55} & \ding{55} & \ding{51} & \ding{51} & \ding{51} \\
        RoboSplat~\cite{robosplat} & \ding{51} & \ding{55} & \ding{51} & \ding{51} & \ding{51} \\
        Real2Render2Real~\cite{yu2025realrenderreal} & \ding{51} & \ding{55} & \ding{51} & \ding{51} & \ding{51} \\
        DemoGen~\cite{XueZ25demogen}  & \ding{51} & \ding{55} & \ding{55} & \ding{55} & \ding{51} \\
        R2RGen~\cite{xu2025r2rgenrealtoreal3ddata} & \ding{51} & \ding{55} & \ding{55} & \ding{55} & \ding{51} \\
        UMIGen~\cite{huang2025umigenunifiedframeworkegocentric} & \ding{51} & \ding{55} & \ding{55} & \ding{55} & \ding{51} \\
        RoboTransfer~\cite{liu2025robotransfer}  & \ding{51} & \ding{51} & \ding{51} & \ding{51} & \ding{55} \\
        MVAug~\cite{tong2025fidelity} & \ding{51} & \ding{51} & \ding{51} & \ding{51} & \ding{55} \\
        \midrule
        Real2Edit2Real (ours) & \ding{51} & \ding{51} & \ding{51} & \ding{51} & \ding{51} \\
        \bottomrule
        \end{tabular}
        %
        \captionof{table}{
            Comparison with Other One-to-many Demonstration Generation Methods.
        }
        \label{tab:supp_compare}
    \end{center}
}]

\section{Contribution Clarification}
To better clarify our contribution, we provide a detailed comparison between our method and other one-to-many demonstration generation approaches, as shown in Table~\ref{tab:supp_compare}. Simulation-based methods like MimicGen~\cite{mandlekar2023mimicgen} and SkillMimicGen~\cite{garrett2024skillmimicgen} rely on simulators and require scene and object assets, which not only introduce a significant sim-to-real gap but also make it difficult to perform data augmentation directly on real-world data. Methods such as RoboSplat~\cite{robosplat} and Real2Render2Real~\cite{yu2025realrenderreal} are built on 3D Gaussian Splatting. Although they do not require a simulation engine, they still rely on dense scanning to reconstruct the objects or scenes. This means that they cannot perform data generation using only the RGB observations from the original demonstrations, which significantly limits their scalability. Another line of research, including DemoGen~\cite{XueZ25demogen}, R2RGen~\cite{xu2025r2rgenrealtoreal3ddata}, and UMIGen~\cite{huang2025umigenunifiedframeworkegocentric} , generates new 3D point-cloud demonstrations through point-cloud editing. However, their reliance on depth sensors limits their compatibility with the current mainstream VLA paradigm that uses multi-view RGB inputs, and also prevents them from performing texture-level augmentation. Methods based on video generation, such as RoboTransfer~\cite{liu2025robotransfer} and MVAug~\cite{tong2025fidelity}, can directly augment multi-view 2D demonstrations, but they only enhance visual aspects such as texture, without increasing the diversity of object spatial distributions or robot trajectories.

In contrast, our method requires no simulator and directly augments the original RGB observations, significantly improving scalability. It simultaneously generates new textures and trajectories for VLA training, highlighting its unified and flexible design.

\section{Real2Edit2Real Implementation Details}
In this section, we provide more details of the proposed framework, Real2Edit2Real:
\begin{itemize}
    \item In Section~\ref{subsec:supp_reconstruction}, we provide additional information for the hybrid training paradigm.
    \item In Section~\ref{subsec:supp_edit}, we explain the full pipeline of depth-reliable spatial editing in detail.
    \item In Section~\ref{subsec:supp_generate}, we discuss more about 3D-controlled video generation model.
\end{itemize}

\subsection{Metric-scale Geometry Reconstruction}
\label{subsec:supp_reconstruction}
\begin{figure*}[t]
    \centering
    \includegraphics[width=\linewidth]{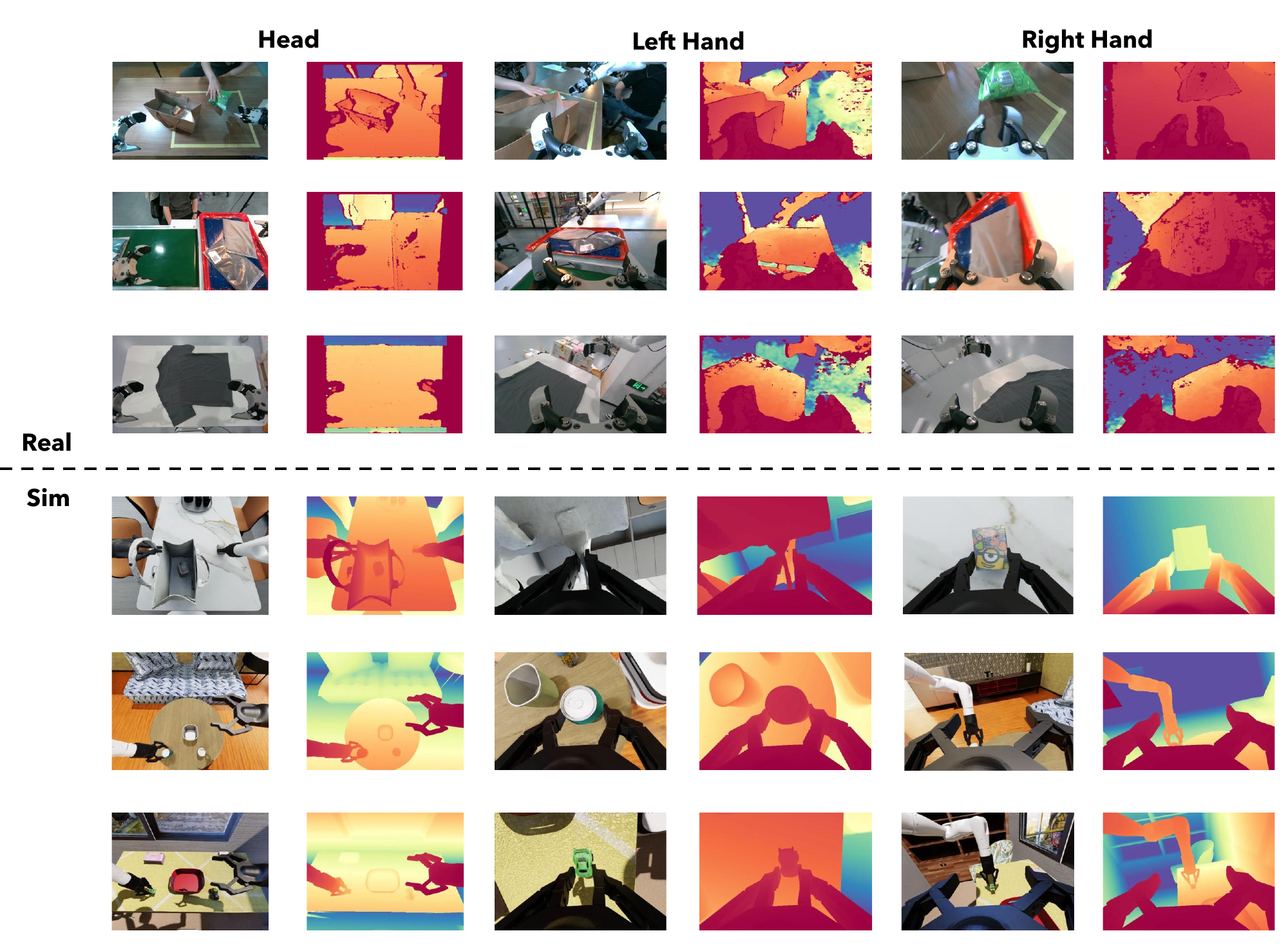}
    \caption{Training Data Visualization of Metric-VGGT. Depth visualization: \textcolor{red}{red} is the nearest, \textcolor{blue}{blue} is the farthest.}
    \label{fig:vggt_traindata_vis}
\end{figure*}

\noindent\textbf{Data Visualization.} Fig.~\ref{fig:vggt_traindata_vis} shows the visualization of the training data. We can see that real-world depth maps are often noisy and contain large invalid regions, whereas synthetic depth is clean and accurate. By training with our proposed hybrid training paradigm, our model learns to reconstruct geometry in metric scale in the real world, effectively compensating for the limitations of depth sensors.

\noindent\textbf{Training Details.}
In Table~\ref{tab:vggt_train_detail}, we provide the details of fine-tuning VGGT~\cite{wang2025vggt} to Metric-VGGT. 

\begin{table}[h]
\renewcommand\tabcolsep{3pt}
\centering
\begin{tabular}{l|l}
\toprule
\textbf{Config} & \textbf{Value} \\
\midrule
Base Model & VGGT-1B \\
Training Real Data & 100,000 \\
Training Sim Data & 40,000 \\
Fine-Tuning Scheme & Full Parameter \\
Total Training Steps & 150,000 \\
Learning Rate & 2e-4 \\
Backbone Learning Rate & 2e-5 \\
LR Scheduler & Cosine Annealing Scheduler \\
ETA Minimum & 1e-6 \\
Weight Decay & 1e-2 \\
View Num & 3 \\
Global Batch Size & 16 \\
Gradient Accumulation Steps & 4 \\
Mixed Precision & bf16 \\
Optimizer & AdamW \\
Training Image Size & 518 \\ 
\bottomrule
\end{tabular}
\caption{Training Details of Metric-VGGT.}
\label{tab:vggt_train_detail}
\end{table}

\subsection{Depth-reliable Spatial Editing}
\label{subsec:supp_edit}

\begin{figure}[h]
    \centering
    \includegraphics[width=\linewidth]{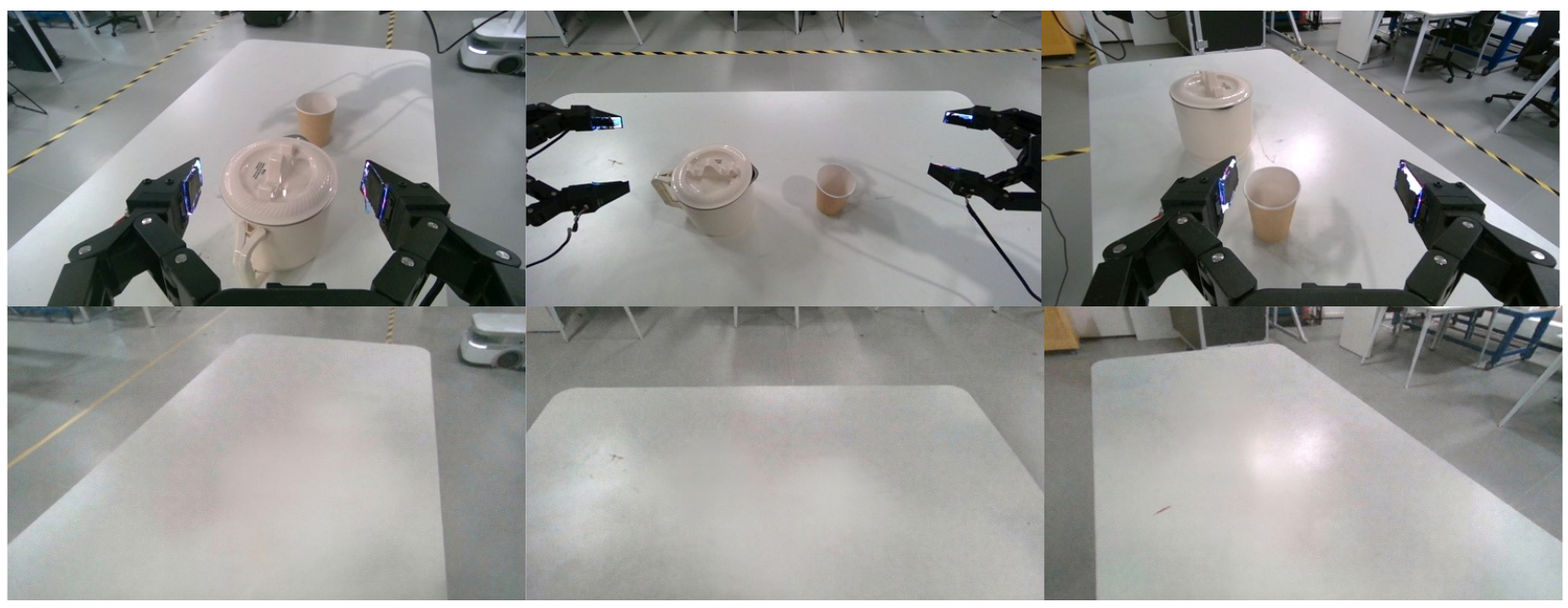}
    \caption{Example of the Inpainted Background.}
    \label{fig:supp_inpaint_example}
\end{figure}

\noindent\textbf{Background Depth Completion.} As we mentioned in the manuscript, projecting edited point clouds to depth maps may cause missing regions in the background due to the object moving and novel robot motion. To mitigate this artifact, we first inpaint the background, which deletes the foreground objects and robot in the multi-view first frames with an image-edit model~\cite{wang2025seededit30fasthighquality}. Figure~\ref{fig:supp_edit_prompt} provides the prompt we used for image editing, and Figure~\ref{fig:supp_inpaint_example} shows an example of the inpainted background. Then, we reconstruct the metric geometry of the background with Metric-VGGT. To correct the metric-scale inconsistencies introduced by image editing, we incorporate an additional point cloud alignment procedure, as shown in Algorithm~\ref{alg:bg_align}.
\begin{figure}[t]
\centering
\begin{minipage}{\linewidth}
\begin{lstlisting}[basicstyle=\ttfamily\footnotesize, breaklines=true]
"While keeping everything else in the image unchanged, remove the black gripper and the black wire."
"While keeping everything else in the image unchanged, remove the white robotic arm."
"While keeping everything else in the image unchanged, remove xxx on the table."
\end{lstlisting}
\end{minipage}
\caption{Prompt Used for Background Inpainting. In the prompt, xxx means the manipulated objects.}
\label{fig:supp_edit_prompt}
\end{figure}

\begin{algorithm}[thbp]
\caption{Background Point-Cloud Alignment}
\label{alg:bg_align}
\begin{algorithmic}
\Require Origin first frame point cloud $\mathcal{P}^{o}$, unaligned background point cloud $\mathcal{P}^{edit}$, table mask $M^{table}$.
\Ensure Metric-aligned background point cloud $\mathcal{P}^{bg}$. 

\Function{EstimatePlane}{$\mathcal{P}$}
    \State $plane \gets \text{RansacPlaneSegment}(\mathcal{P})$
    \State // $plane: [a, b, c, d], ax + by + cz + d = 0$
    \State \Return $plane$ 
\EndFunction

\State $plane^{o} \gets \text{EstimatePlane}(\mathcal{P}^o[M^{table}])$
\State $plane^{edit} \gets \text{EstimatePlane}(\mathcal{P}^{edit}[M^{table}])$
\State $scale \gets plane^o[3] / plane^{edit}[3]$
\State $\mathcal{P}^{bg} \gets scale \times \mathcal{P}^{edit}$
\State \Return $\mathcal{P}^{bg}$
\end{algorithmic}
\end{algorithm}

\noindent\textbf{Spatial Editing Pipeline.} After getting a completed background point cloud, we separate foreground objects through Grounded-SAM~\cite{kirillov2023segany, liu2023grounding, ren2024grounded} and robotic dual arms through forward kinematics. Following, we provide an example to detail the full spatial editing pipeline. Algorithm~\ref{alg:supp_edit_mug2basket}, \ref{alg:supp_edit_mug2basket-2} shows the spatial editing pipeline of the Mug to Basket task. The Object Relocation Segment produces the depth sequence for Smooth Object Relocation described in the manuscript.

\begin{algorithm}[t]
\caption{Pipeline of Mug to Basket}
\label{alg:supp_edit_mug2basket}
\begin{algorithmic}
\Require Source point clouds $\mathcal{P}^{l}$, $\mathcal{P}^{r}$, $\mathcal{P}^{mug}$, $\mathcal{P}^{basket}$, background point cloud $\mathcal{P}^{bg}$, joint states $\mathcal{Q}$, action trajectory $\mathcal{A}$, camera poses $\mathcal{T}^{h}$, $\mathcal{T}^{l}$, $\mathcal{T}^{r}$, skill-1 start timestep $t_1$, skill-1 end timestep $t_2$, skill-2 start timestep $t_3$, skill-2 end timestep $t_4$.
\Ensure Novel depth sequence $\mathcal{D}^{h\star}$, $\mathcal{D}^{l\star}$, $\mathcal{D}^{r\star}$, joint states $\mathcal{Q}^\star$, action trajectory $\mathcal{A}^\star$, camera poses $\mathcal{T}^\star$. 

\Function{RenderDepth}{$\mathcal{P}, \mathcal{T}, \mathcal{Q}$}
    \State $D_1 \gets \text{ProjectPointCloud}(\mathcal{P}, \mathcal{T})$
    \State $D_2 \gets \text{RenderLinkDepth}(\mathcal{Q}, \mathcal{T})$
    \State \Return $\text{Merge}(D_1, D_2)$
\EndFunction
\State Sample a Object Transform Pair $\mathbf{T}^{mug},\mathbf{T}^{basket} \in \mathbb{R}^{4\times 4}$
\State $\mathcal{D}^{h\star} \gets list()$, 
$\mathcal{D}^{l\star} \gets list()$, 
$\mathcal{D}^{r\star} \gets list()$
\State $\mathcal{Q}^\star \gets list()$, 
$\mathcal{A}^\star \gets list()$, 
$\mathcal{T}^\star \gets list()$ 
\State // Object Relocation Segment
\For{$t$ in range(0, 30)}
    \State $\mathbf{T}^{mug}_t,\mathbf{T}^{basket}_t \gets \text{Interpolate}(\mathbf{T}^{mug},\mathbf{T}^{basket}, 30, t)$
    \State $\mathcal{P}^\star_t \gets \mathbf{T}^{mug}_t\mathcal{P}^{mug}_0 \cup \mathbf{T}^{basket}_t\mathcal{P}^{basket}_0 \cup \mathcal{P}^{l}_0 \cup \mathcal{P}^{r}_0 \cup \mathcal{P}^{bg}$
    \State $\mathcal{D}^{h*} \gets \mathcal{D}^{h*} \cup \text{RenderDepth}(\mathcal{P}^\star_t, \mathcal{T}^{h}_0, \mathcal{Q}_0)$
    \State $\mathcal{D}^{l*} \gets \mathcal{D}^{l*} \cup \text{RenderDepth}(\mathcal{P}^\star_t, \mathcal{T}^{l}_0, \mathcal{Q}_0)$
    \State $\mathcal{D}^{r*} \gets \mathcal{D}^{r*} \cup \text{RenderDepth}(\mathcal{P}^\star_t, \mathcal{T}^{r}_0, \mathcal{Q}_0)$
    \State $\mathcal{Q^\star} \gets  \mathcal{Q}^\star \cup \mathcal{Q}_0$, $\mathcal{A}^\star \gets \mathcal{A}^\star \cup \mathcal{A}_0$
    \State $\mathcal{T}^\star \gets \mathcal{T}^\star \cup (\mathcal{T}^{h}_0, \mathcal{T}^{l}_0, \mathcal{T}^{r}_0)$
\EndFor
\State // Motion-1 Segment
\State $\mathcal{A}^\star_{start} \gets \mathcal{A}_0$, 
$\mathcal{A}^\star_{end} \gets \mathbf{T}^{mug}A_{t_1}$,
\For{$t$ in range(0, $t_1$)}
    \State $\mathbf{T}_t, \mathcal{A}^\star_t,\mathcal{Q}^\star_t \gets \text{MotionPlan}(\mathcal{A}^\star_{start}, \mathcal{A}^\star_{end}, t)$
    \State $\mathcal{P}^{ree}_t \gets \mathcal{P}^{r}_t \setminus \text{FK}(\mathcal{P}^{r}_t, \mathcal{Q}_t)$
    \State $\mathcal{P}^\star_t \gets \mathbf{T}_t\mathcal{P}^{ree}_t \cup 
    \mathcal{P}^{l}_t \cup 
    \mathbf{T}^{mug}\mathcal{P}^{mug}_t \cup \mathbf{T}^{basket}\mathcal{P}^{basket}_t \cup \mathcal{P}_{bg}$
    \State $\mathcal{D}^{h\star} \gets \mathcal{D}^{h\star} \cup \text{RenderDepth}(\mathcal{P}^\star_t, \mathcal{T}^{h}_t, \mathcal{Q}^\star_t)$
    \State $\mathcal{D}^{l\star} \gets \mathcal{D}^{l\star} \cup \text{RenderDepth}(\mathcal{P}^\star_t, \mathcal{T}^{l}_t, \mathcal{Q}^\star_t)$
    \State $\mathcal{D}^{r\star} \gets \mathcal{D}^{r\star} \cup \text{RenderDepth}(\mathcal{P}^\star_t, \mathbf{T}_t\mathcal{T}^{r}_t, \mathcal{Q}^\star_t)$
    \State $\mathcal{Q^\star} \gets  \mathcal{Q^\star} \cup \mathcal{Q}^\star_t$, $\mathcal{A}^\star \gets \mathcal{A}^\star \cup \mathcal{A}^\star_t$
    \State $\mathcal{T}^\star \gets \mathcal{T}^\star \cup (\mathcal{T}^h_t, \mathcal{T}^l_t, \mathbf{T}_t\mathcal{T}^r_t$)
\EndFor
\State // Skill-1 Segment
\For{$t$ in range($t_1$, $t_2$)}
    \State $\mathcal{Q}^\star_t \gets \text{IK}(\mathbf{T}^{mug}\mathcal{A}_t)$
    \State $\mathcal{P}^{ree}_t \gets \mathcal{P}^{r}_t \setminus \text{FK}(\mathcal{P}^{r}_t, \mathcal{Q}_t)$
    \State $\mathcal{P}^\star_t \gets \mathbf{T}^{mug}(\mathcal{P}^{ree}_t \cup \mathcal{P}^{mug}_t) \cup \mathcal{P}^l_t \cup \mathbf{T}^{mug}\mathcal{P}^{mug}_t \cup \mathcal{P}_{bg}$
    \State $\mathcal{D}^{h\star} \gets \mathcal{D}^{h\star} \cup \text{RenderDepth}(\mathcal{P}^\star_t, \mathcal{T}^{h}_t, \mathcal{Q}^\star_t)$
    \State $\mathcal{D}^{l\star} \gets \mathcal{D}^{l\star} \cup \text{RenderDepth}(\mathcal{P}^\star_t, \mathcal{T}^{l}_t, \mathcal{Q}^\star_t)$
    \State $\mathcal{D}^{r\star} \gets \mathcal{D}^{r\star} \cup \text{RenderDepth}(\mathcal{P}^\star_t, \mathbf{T}^{mug}\mathcal{T}^{r}_t, \mathcal{Q}^\star_t)$
    \State $\mathcal{Q^\star} \gets  \mathcal{Q^\star} \cup \mathcal{Q}^\star_t$, $\mathcal{A}^\star \gets \mathcal{A}^\star \cup \mathbf{T}^{mug}\mathcal{A}_t$ 
    \State $\mathcal{T}^\star \gets \mathcal{T}^\star \cup (\mathcal{T}^h_t, \mathcal{T}^l_t, \mathbf{T}^{mug}\mathcal{T}^r_t$)
\EndFor
\end{algorithmic}
\end{algorithm}

\begin{algorithm}[t]
\caption{Continued to Pipeline of Mug to Basket}
\label{alg:supp_edit_mug2basket-2}
\begin{algorithmic}

\State // Motion-2 Segment
\State $\mathcal{A}^\star_{start} \gets \mathbf{T}^{mug}\mathcal{A}_{t_2}$, 
$\mathcal{A}^\star_{end} \gets \mathbf{T}^{basket}A_{t_3}$,
\For{$t$ in range($t_2$, $t_3$)}
    \State $\mathbf{T}_t, \mathcal{A}^\star_t,\mathcal{Q}^\star_t \gets \text{MotionPlan}(\mathcal{A}^\star_{start}, \mathcal{A}^\star_{end}, t)$
    \State $\mathcal{P}^{ree}_t \gets \mathcal{P}^{r}_t \setminus \text{FK}(\mathcal{P}^{r}_t, \mathcal{Q}_t)$
    \State $\mathcal{P}^\star_t \gets \mathbf{T}_t(\mathcal{P}^{ree}_t \cup \mathcal{P}^{mug}_t) \cup
    \mathcal{P}^{l}_t \cup 
    \mathbf{T}^{basket}\mathcal{P}^{basket}_t \cup \mathcal{P}_{bg}$
    \State $\mathcal{D}^{h\star} \gets \mathcal{D}^{h\star} \cup \text{RenderDepth}(\mathcal{P}^\star_t, \mathcal{T}^{h}_t, \mathcal{Q}^\star_t)$
    \State $\mathcal{D}^{l\star} \gets \mathcal{D}^{l\star} \cup \text{RenderDepth}(\mathcal{P}^\star_t, \mathcal{T}^{l}_t, \mathcal{Q}^\star_t)$
    \State $\mathcal{D}^{r\star} \gets \mathcal{D}^{r\star} \cup \text{RenderDepth}(\mathcal{P}^\star_t, \mathbf{T}_t\mathcal{T}^{r}_t, \mathcal{Q}^\star_t)$
    \State $\mathcal{Q^\star} \gets  \mathcal{Q^\star} \cup \mathcal{Q}^\star_t$, $\mathcal{A}^\star \gets \mathcal{A}^\star \cup \mathcal{A}^\star_t$
    \State $\mathcal{T}^\star \gets \mathcal{T}^\star \cup (\mathcal{T}^h_t, \mathcal{T}^l_t, \mathbf{T}_t\mathcal{T}^r_t$)
\EndFor
\State // Skill-2 Segment
\For{$t$ in range($t_3$, $t_4$)}
    \State $\mathcal{Q}^\star_t \gets \text{IK}(\mathbf{T}^{basket}\mathcal{A}_t)$
    \State $\mathcal{P}^{ree}_t \gets \mathcal{P}^{r}_t \setminus \text{FK}(\mathcal{P}^{r}_t, \mathcal{Q}_t)$
    \State $\mathcal{P}^\star_t \gets \mathbf{T}^{basket}(\mathcal{P}^{ree}_t \cup \mathcal{P}^{mug}_t \cup \mathcal{P}^{basket}_t) \cup \mathcal{P}^l_t \cup \mathcal{P}_{bg}$
    \State $\mathcal{D}^{h\star} \gets \mathcal{D}^{h\star} \cup \text{RenderDepth}(\mathcal{P}^\star_t, \mathcal{T}^{h}_t, \mathcal{Q}^\star_t)$
    \State $\mathcal{D}^{l\star} \gets \mathcal{D}^{l\star} \cup \text{RenderDepth}(\mathcal{P}^\star_t, \mathcal{T}^{l}_t, \mathcal{Q}^\star_t)$
    \State $\mathcal{D}^{r\star} \gets \mathcal{D}^{r\star} \cup \text{RenderDepth}(\mathcal{P}^\star_t, \mathbf{T}^{basket}\mathcal{T}^{r}_t, \mathcal{Q}^\star_t)$
    \State $\mathcal{Q^\star} \gets  \mathcal{Q^\star} \cup \mathcal{Q}^\star_t$, $\mathcal{A}^\star \gets \mathcal{A}^\star \cup \mathbf{T}^{basket}\mathcal{A}_t$ 
    \State $\mathcal{T}^\star \gets \mathcal{T}^\star \cup (\mathcal{T}^h_t, \mathcal{T}^l_t, \mathbf{T}^{baseket}\mathcal{T}^r_t$)
\EndFor

\State \Return $\mathcal{D}^\star$, $\mathcal{Q}^\star$, $\mathcal{A}^\star$, $\mathcal{T}^\star$ 
\end{algorithmic}
\end{algorithm}

\subsection{3D-Controlled Video Generation}
\label{subsec:supp_generate}
\noindent\textbf{Training Data.} For training the 3D-controlled multi-view video generation model, we sample 7K episodes of 64 tasks from the Agibot-World datasets~\cite{agibot2025agibotworld}. To get the control conditions of the training data, we used the Metric-VGGT to predict the depth maps and compute the Canny Edges from depth. To ensure the 3D control condition remains consistent across multi-view and temporal, we perform global normalization on the depth sequences of all three views within a training chunk, rather than normalizing each depth map individually.

\noindent\textbf{Condition Dropout.} In the training stage, we fine-tune the backbone of GE-Sim~\cite{liao2025genie} (based on Cosmos-Predict-2B~\cite{contributors2025agibotdigitalworld}) with sampling data from the Agibot-World Dataset~\cite{agibot2025agibotworld}. In multi-condition compositional generation, intensity-based conditions such as depth maps and Canny edges tend to dominate the visual information, potentially diminishing the influence of other control signals during training~\cite{huang2023composercreativecontrollableimage}. However, these two conditions always introduce noise after spatial editing. To improve robustness against imperfect control signals, we apply random dropout to the depth and Canny edge conditions during training, where they are independently dropped with a probability of 0.5, and jointly dropped with a probability of 0.1.
By randomly masking portions of these inputs, the model is encouraged to rely on complementary visual evidence, rather than depending solely on the intensity conditions, ultimately improving the realism of the generated videos under noisy conditions.

\noindent\textbf{Training Details.}
In Table~\ref{tab:generate_train_detail}, we provide the details of training the 3D-controlled video generation model.

\begin{table}[t]
\renewcommand\tabcolsep{4pt}
\centering
\begin{tabular}{l|l}
\toprule
\textbf{Config} & \textbf{Value} \\
\midrule
Base Model & GE-Sim-2B \\
Training Data & 7000 Episodes 64 Tasks\\
Fine-Tuning Scheme & Full Parameter \\
Total Training Steps & 20,000 \\
Learning Rate & 1e-4 \\
LR Scheduler & Constant with Warmup \\
LR Warmup Steps & 1000 \\
Weight Decay & 5e-5 \\
Global Batch Size & 16 \\
Gradient Accumulation Steps & 1 \\
Max Gradient Norm & 1.0 \\
Mixed Precision & bf16 \\
Optimizer & AdamW \\
Training Resolution & 384$\times$512 \\ 
Video Chunk Length & 25 \\
Memory Frames & 4 \\
\bottomrule
\end{tabular}
\caption{Training Details of 3D-controlled Video Generation Model.}
\label{tab:generate_train_detail}
\end{table}

\section{Experiment Details}

\begin{figure}[t]
    \centering
    \includegraphics[width=\linewidth]{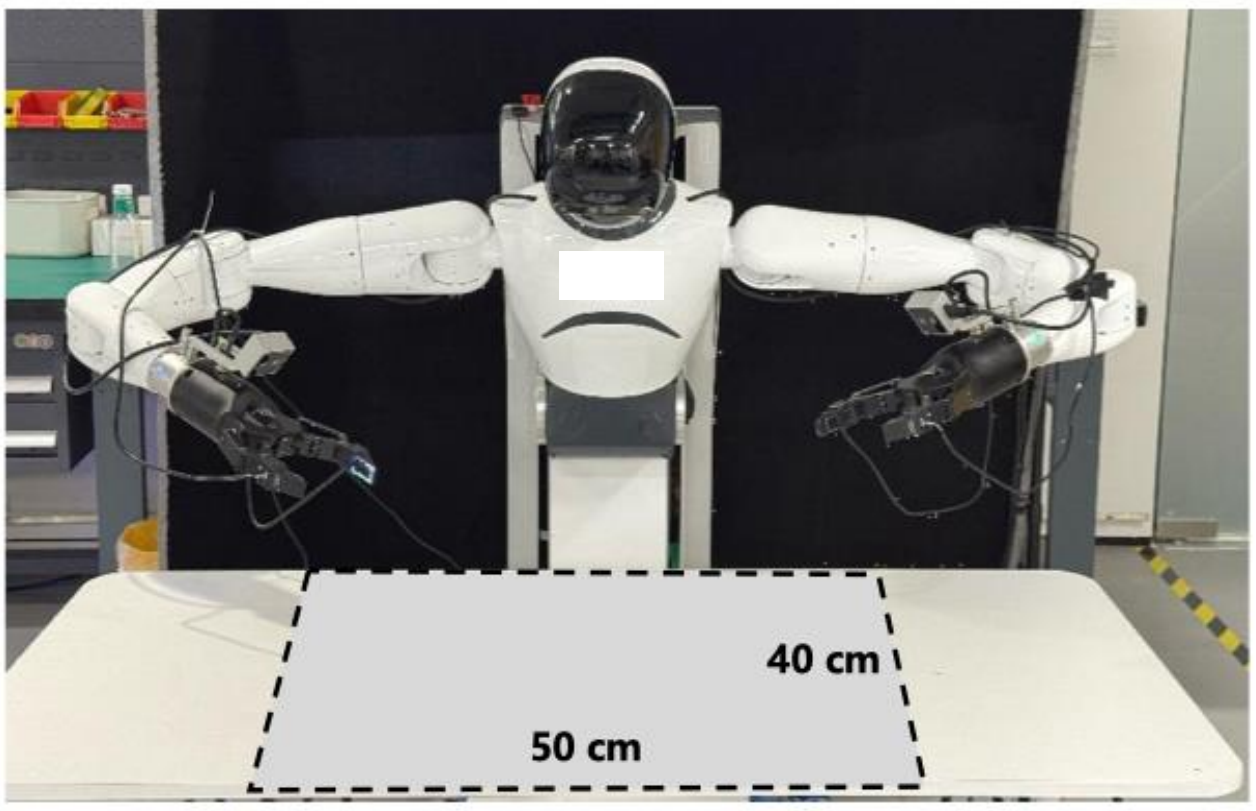}
    \caption{Visualization of Manipulation Workspace.}
    \label{fig:workspace}
\end{figure}

\textbf{Workspace.} Figure~\ref{fig:workspace} shows the workspace of four real-robot manipulation tasks in the manuscript. The workspace is determined by the maximal range in which the robot’s kinematic configuration can perform the intended tasks.

\begin{figure}[t]
    \centering
    \includegraphics[width=\linewidth]{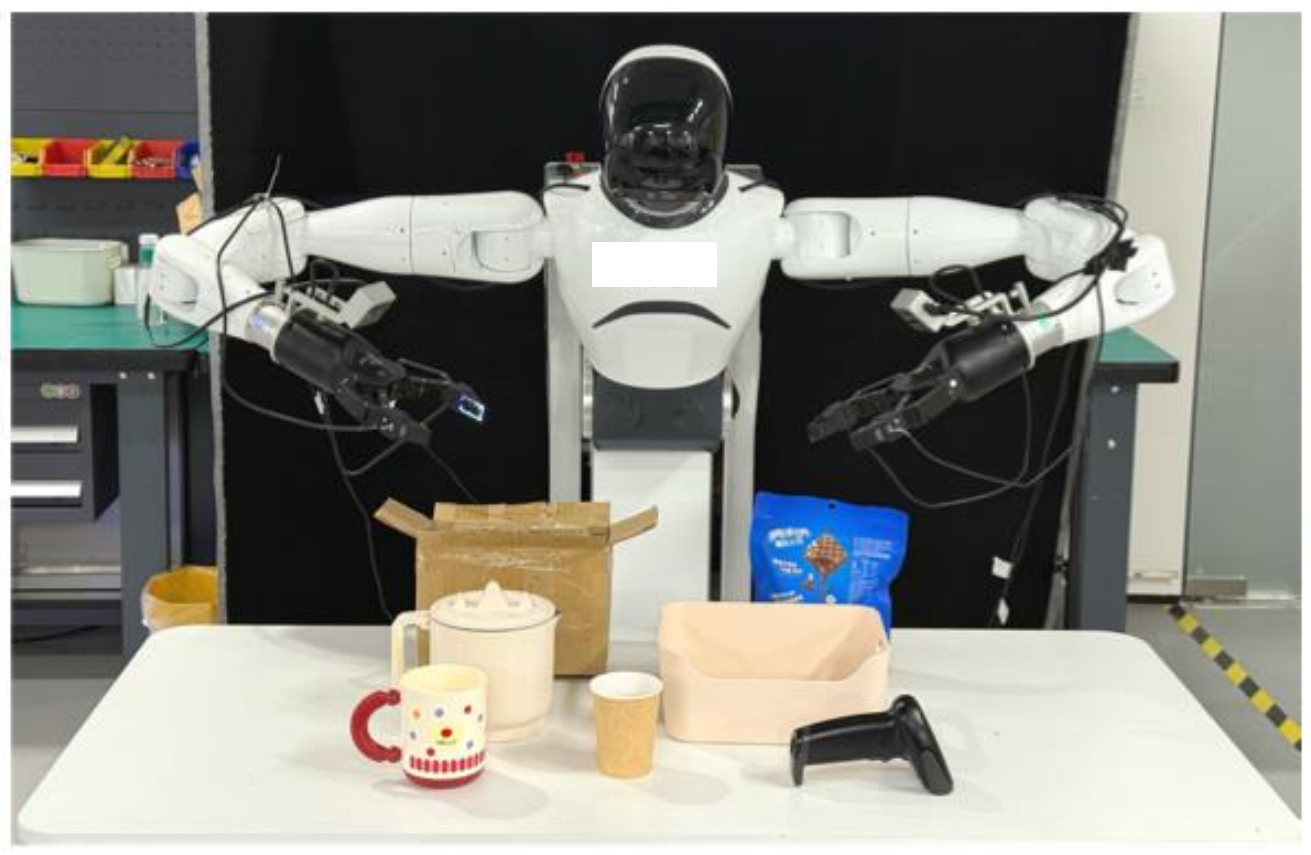}
    \caption{Visualization of Manipulation Objects.}
    \label{fig:object}
\end{figure}

\noindent\textbf{Object Set.} Figure~\ref{fig:object} shows all the objects we used in the manipulation tasks. Because it is impractical to verify every object in the training set, all objects in the figure are newly purchased to minimize any potential overlap with the dataset. This setup enables a more reliable assessment of our framework’s generalization to unseen objects. In addition, the real-world testing laboratory is also absent from the training data used to develop our framework.

\noindent\textbf{Data Generation.} To generate demonstrations, we first reconstruct the source demonstrations and apply a confidence threshold between 30\% and 50\% to remove spurious points. During spatial editing, we define an augmentation region around the object’s original location, typically a 40cm$\times$40cm square, and augment object rotations within a 30°–60° range. For video generation, we set the diffusion step to 6 and the memory length to 4, for which we uniformly sample from the already generated frames, including both the first and last frames.

\section{Additional Experiments}

\subsection{Quantitative Results of Video Generation}
To evaluate the performance of our video generation model, we conduct experiments using data collected from the four real-world robot tasks to prevent data contamination. Table~\ref{tab:video_gen_results} shows the quantitative results compared to GE-Sim~\cite{liao2025genie} with the conditional I2V setting. It demonstrates that our video generation module produces robot demonstrations with significantly enhanced visual realism.

\begin{table}[h]
\centering
\begin{tabular}{lcccc}
\toprule
Method & FVD $\downarrow$ & LPIPS $\downarrow$ & SSIM $\uparrow$ & PSNR $\uparrow$ \\
\midrule
GE-Sim~\cite{liao2025genie} & 663.4 & 0.2038 & 0.7491 & 20.41 \\
\textbf{Ours} & \textbf{352.9} & \textbf{0.1252} & \textbf{0.8647} & \textbf{22.95} \\
\bottomrule
\end{tabular}
\caption{Quantitative Results of Video Generation. We compare our method with GE-Sim across several standard metrics on conditional I2V. Bold numbers indicate the best performance.}
\label{tab:video_gen_results}
\end{table}

\subsection{Diffusion Policy on Mug to Basket}
To further validate the quality of the data generated by Real2Edit2Real, we conduct additional Diffusion Policy~\cite{chi2023diffusionpolicy} experiments on the Mug to Basket task. We use a ViT-S encoder initialized with DINO-v3~\cite{siméoni2025dinov3} weights and train the Diffusion Policy in a full-parameter manner on different training data. Table~\ref{tab:dp_mug2basket} shows the success rate of diffusion polices trained with real demonstrations and generated demonstrations, which indicates that generating data from only a few source demonstrations, like 1-5, can make DP surpass that trained with 50 real demonstrations on this task.

\begin{table}[h]
\renewcommand\tabcolsep{6pt}
\centering
\begin{tabular}{cccccc}
\toprule
\textbf{R10} & \textbf{R20} & \textbf{R50} & \textbf{R1G200} & \textbf{R2G200} & \textbf{R5G200} \\
\midrule
0/20 & 9/20 & 11/20 & 13/20 & 14/20 & 17/20\\
\bottomrule
\end{tabular}
\caption{Performance of Diffusion Policy on the Mug to Basket task. R means the number of real demonstrations, and G means the number of generated demonstrations.}
\label{tab:dp_mug2basket}
\end{table}

\subsection{Generation Time Analysis}
\label{sec:speed}

\begin{figure}[h]
    \centering
    \includegraphics[width=\linewidth]{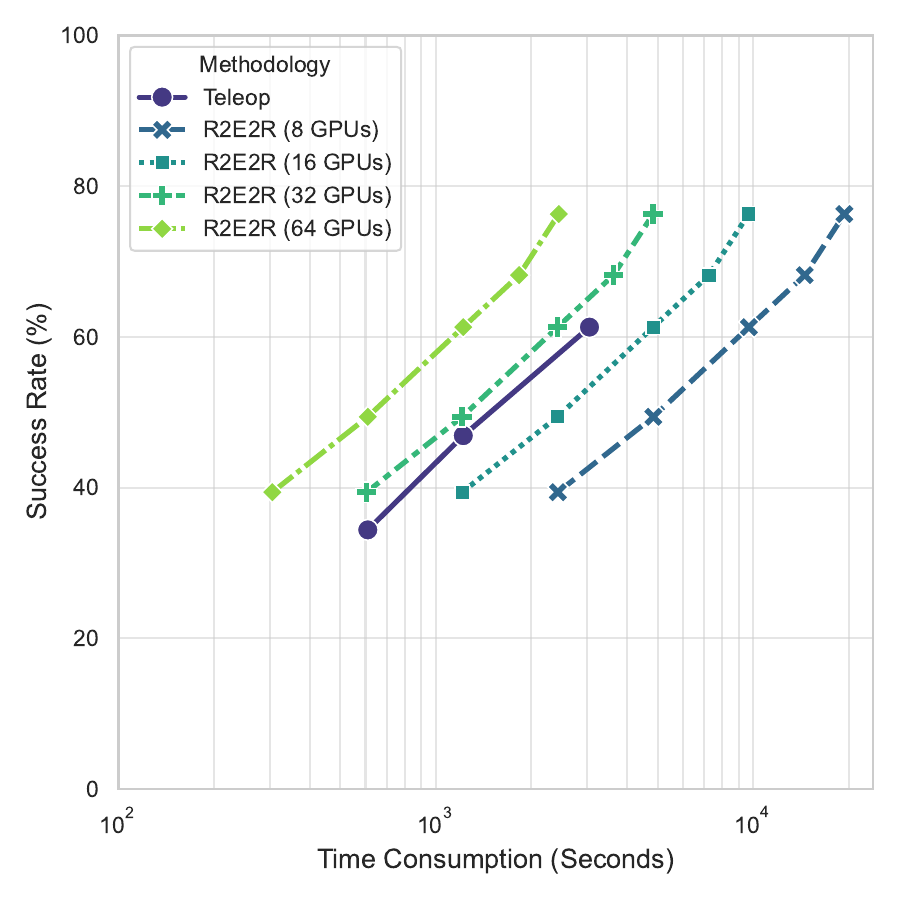}
    \caption{Time Analysis of Real2Edit2Real. We report the success rate (\%) relative to the total time consumption (seconds, log-scale) for data generation. }
    \label{fig:speed}
\end{figure}

Figure~\ref{fig:speed} presents the generation time analysis of the Real2Edit2Real framework. While the integration of video generation modules introduces a computational bottleneck at lower GPU counts, our approach exhibits parallel scalability. By leveraging multi-GPU acceleration, the data generation time is significantly reduced, allowing Real2Edit2Real to surpass the success rates of manual teleoperation in a short time.

\subsection{Generation Data Scaling Analysis}
\begin{figure}[t]
    \centering
    \includegraphics[width=\linewidth]{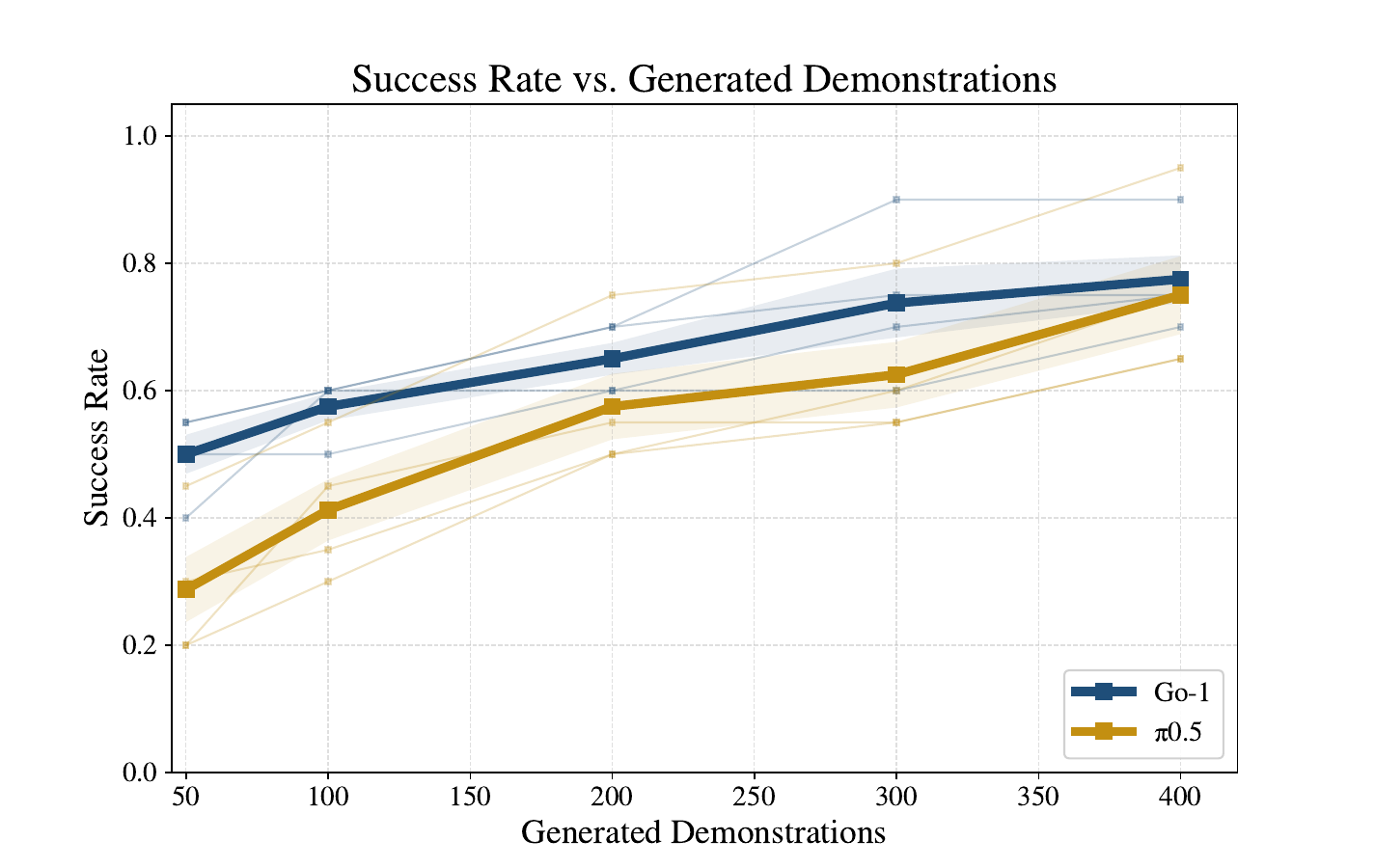}
    \caption{Scaling Analysis of Generated Demonstrations. Bold curves denote the task-averaged performance, while the faint translucent curves visualize the trajectories of individual tasks. Both policies exhibit consistently improved success rates when scaling up generated demonstrations.}
    \vspace{-15pt}
    \label{fig:gen_scaling}
\end{figure}

In the manuscript, we investigate how increasing the number of source demonstrations affects policy performances. Here, we additionally examine, within our proposed Real2Edit2Real framework, the impact of generating more demonstrations. To this end, we produce varying numbers of demonstrations from a single source demonstration and evaluate the resulting policies, using the same training and evaluation protocols as in the manuscript. Experimental results are shown in Figure~\ref{fig:gen_scaling} and Table~\ref{tab:gen_scaling}. The results indicate: (1) Both policies exhibit consistently improved success rates when scaling up generated demonstrations. (2) When we generate more than 300 demonstrations from only one demo, the average success
rates surpass that of 50 real demonstrations.

\begin{table*}[t]
\renewcommand\tabcolsep{7pt}
\centering
\begin{tabular}{l|cccccccc|cc}
\toprule
\multirow{2}{*}{\textbf{\# Demo}} & \multicolumn{2}{c}{\textbf{Mug to Basket}} & \multicolumn{2}{c}{\textbf{Pour Water}} & \multicolumn{2}{c}{\textbf{Lift Box}} & \multicolumn{2}{c}{\textbf{Scan Barcode}} & \multicolumn{2}{|c}{\textbf{Total}}\\
\cmidrule(lr){2-11}
& Go-1 & $\pi_{0.5}$ & Go-1 & $\pi_{0.5}$ & Go-1 & $\pi_{0.5}$ & Go-1 & $\pi_{0.5}$ &  Go-1 & $\pi_{0.5}$ \\
\midrule
Real 10 & 8 / 20 & 8 / 20 & 5 / 20 & 1 / 20 & 11 / 20 & 13 / 20 & 5 / 20 & 4 / 20 & 36.3\% & 32.5\%\\
Real 20 & 12 / 20 & 14 / 20 & 7 / 20 & 2 / 20 & 12 / 20 & 15 / 20 & 8 / 20 & 5 / 20 & 48.8\% & 45.0\%\\
Real 50 & 14 / 20 & 13 / 20 & 8 / 20 & 8 / 20 & 15 / 20 & 17 / 20 & 12 / 20 & 11 / 20 & 61.3\% & 61.3\%\\
\midrule
Real 1 Gen 50 & 8 / 20 & 9 / 20 & 11 / 20 & 4 / 20 & 10 / 20 & 6 / 20 & 11 / 20 & 4 / 20 & 50.0\% & 28.8\% \\
Real 1 Gen 100 & 12 / 20 & 11 / 20 & 12 / 20 & 6 / 20 & 10 / 20 & 7 / 20 & 12 / 20 & 9 / 20 & 57.5\% & 41.3\% \\
Real 1 Gen 200 & 14 / 20 & 15 / 20 & 12 / 20 & 10 / 20 & 12 / 20 & 10 / 20 & 14 / 20 & 11 / 20 & 65.0\% & 57.5\% \\
Real 1 Gen 300 & 15 / 20 & 16 / 20 & 12 / 20 & 12 / 20 & 14 / 20 & 11 / 20 & 18 / 20 & 11 / 20 & 73.8\% & 62.5\%\\
Real 1 Gen 400 & 15 / 20 & 19 / 20 & 14 / 20 & 15 / 20 & 15 / 20 & 13 / 20 & 18 / 20 & 13 / 20 & 77.5\% & 75.0\%\\
\bottomrule
\end{tabular}
\caption{Scaling Analysis of Generated Demonstrations. This compares the performance of polices trained with different numbers of demonstrations generated from only one source demonstration. We can see that increasing the number of generated demonstrations leads to improved success rates for both policies. When we generate more than 300 demonstrations from only one demo, the average success rates even surpass that of 50 real demonstrations.}
\label{tab:gen_scaling}
\vspace{-10pt}
\end{table*}

\subsection{Ablation Study of Control Conditions}

\begin{figure*}[t]
    \centering
    \includegraphics[width=\textwidth]{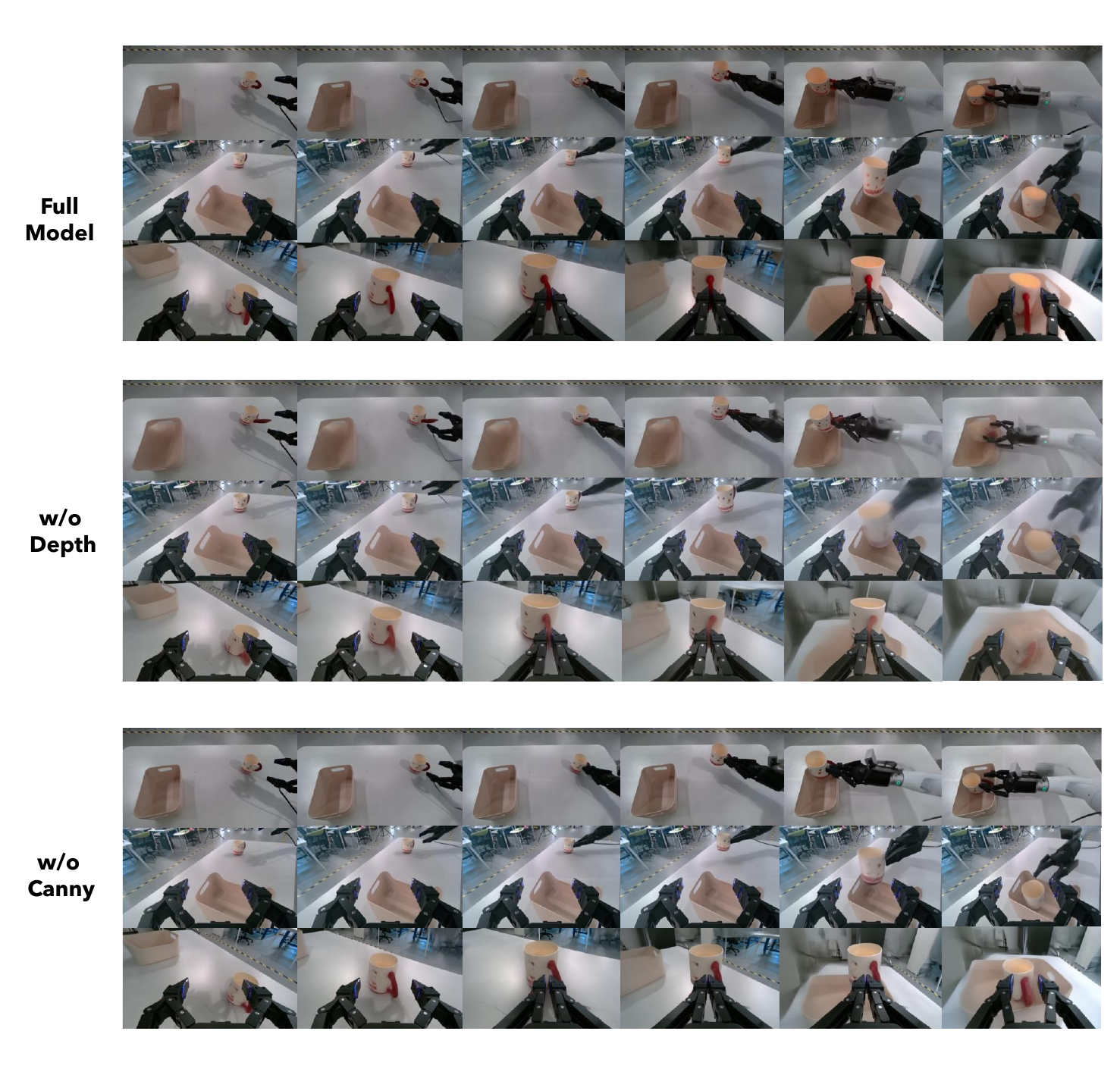}
    \caption{Ablation Study of Control Conditions on Mug to Basket.}
    \label{fig:supp_ablation_mug}
\end{figure*}

\begin{figure*}[t]
    \centering
    \includegraphics[width=\textwidth]{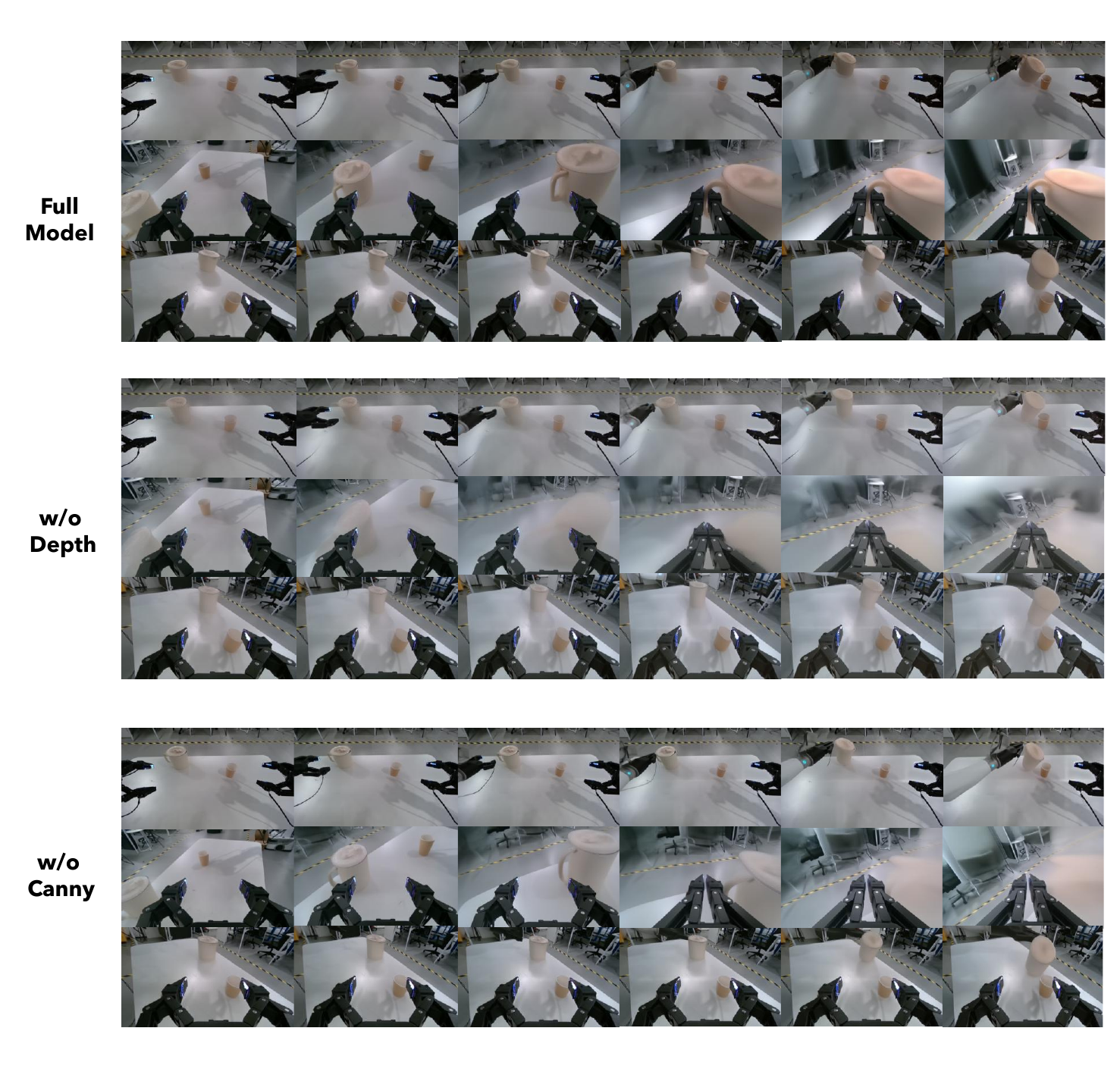}
    \caption{Ablation Study of Control Conditions on Pour Water.}
    \label{fig:supp_ablation_pour}
\end{figure*}

\begin{figure*}[t]
    \centering
    \includegraphics[width=\textwidth]{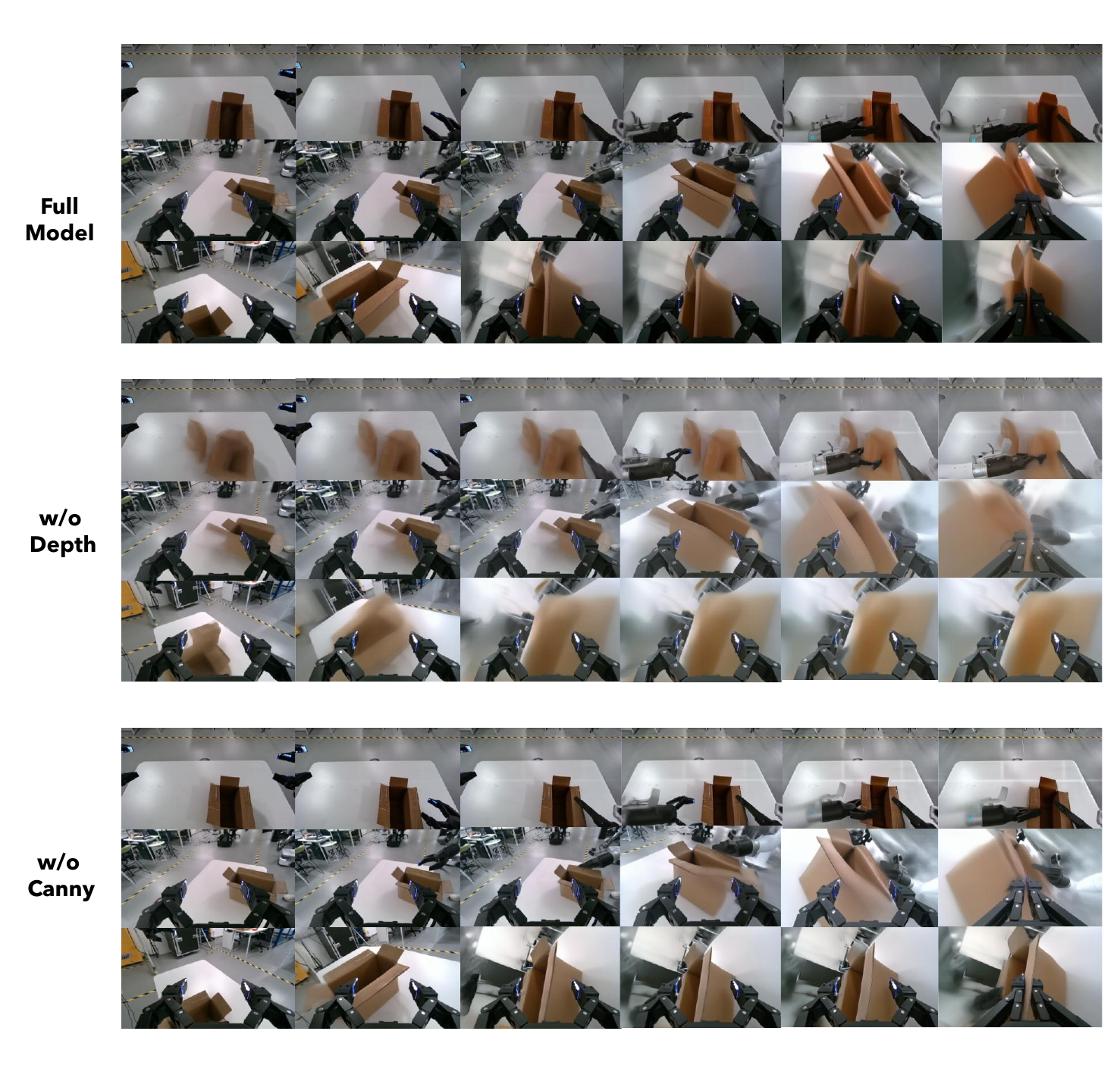}
    \caption{ Ablation Study of Control Conditions on Lift Box.}
    \label{fig:supp_ablation_lift}
\end{figure*}

\begin{figure*}[t]
    \centering
    \includegraphics[width=\textwidth]{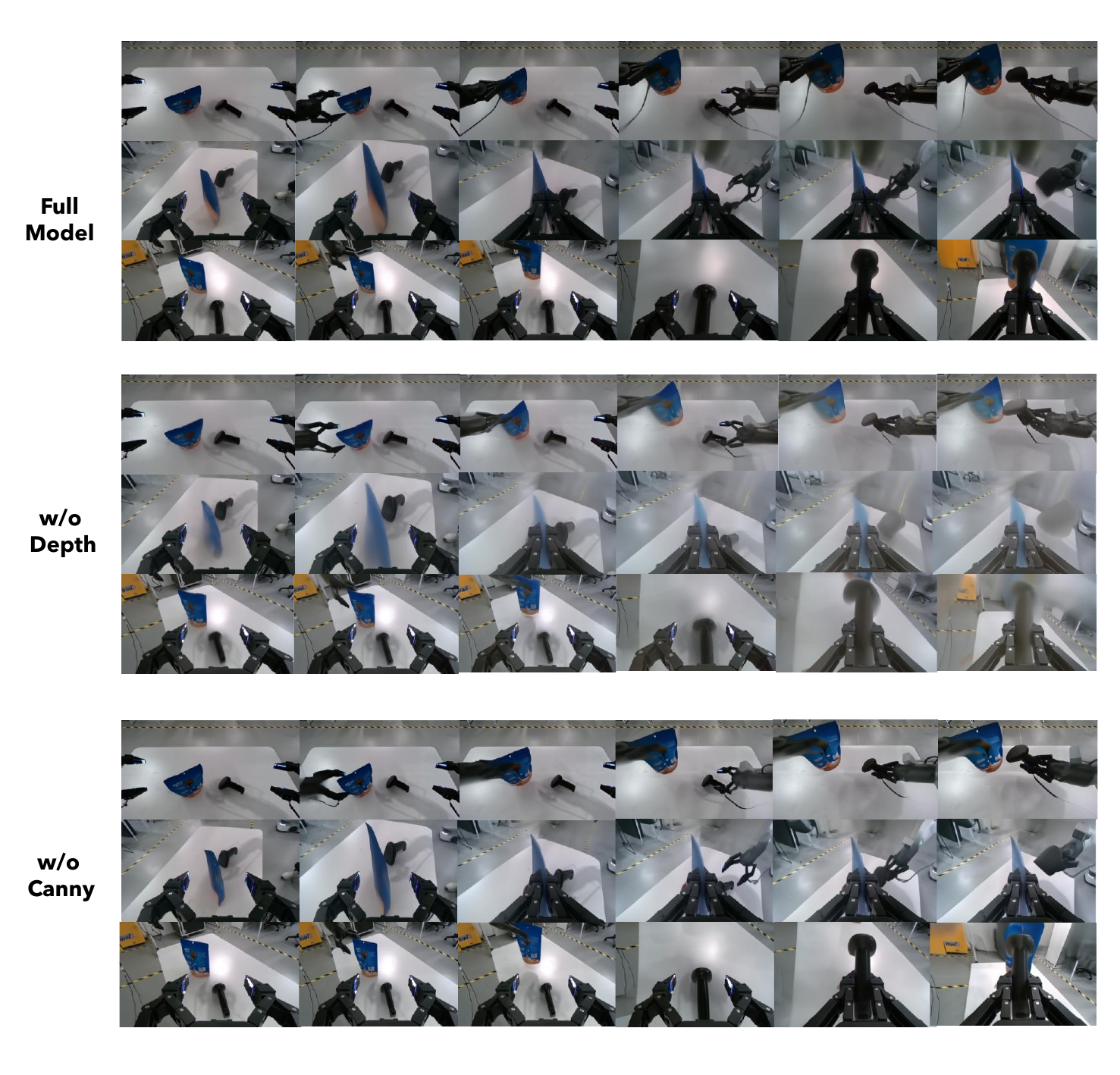}
    \caption{Ablation Study of Control Conditions on Scan Barcode.}
    \label{fig:supp_ablation_scan}
\end{figure*}

In the manuscript, we introduced our 3D-controlled video generation model, which uses depth as the 3D control interface and incorporates Canny edges computed from the depth map as an auxiliary condition. To investigate the roles of depth and Canny edges in video generation, we conduct qualitative ablation studies by removing each condition individually. Fig.~\ref{fig:supp_ablation_mug}, \ref{fig:supp_ablation_pour}, \ref{fig:supp_ablation_lift}, \ref{fig:supp_ablation_scan} show the results on four tasks, respectively. The results demonstrate that removing either the depth control or the Canny edge constraint leads to issues such as object blurring and incorrect interactions, which substantially degrade the quality of the generated demonstrations.

\begin{figure}[h]
    \centering
    \includegraphics[width=\linewidth]{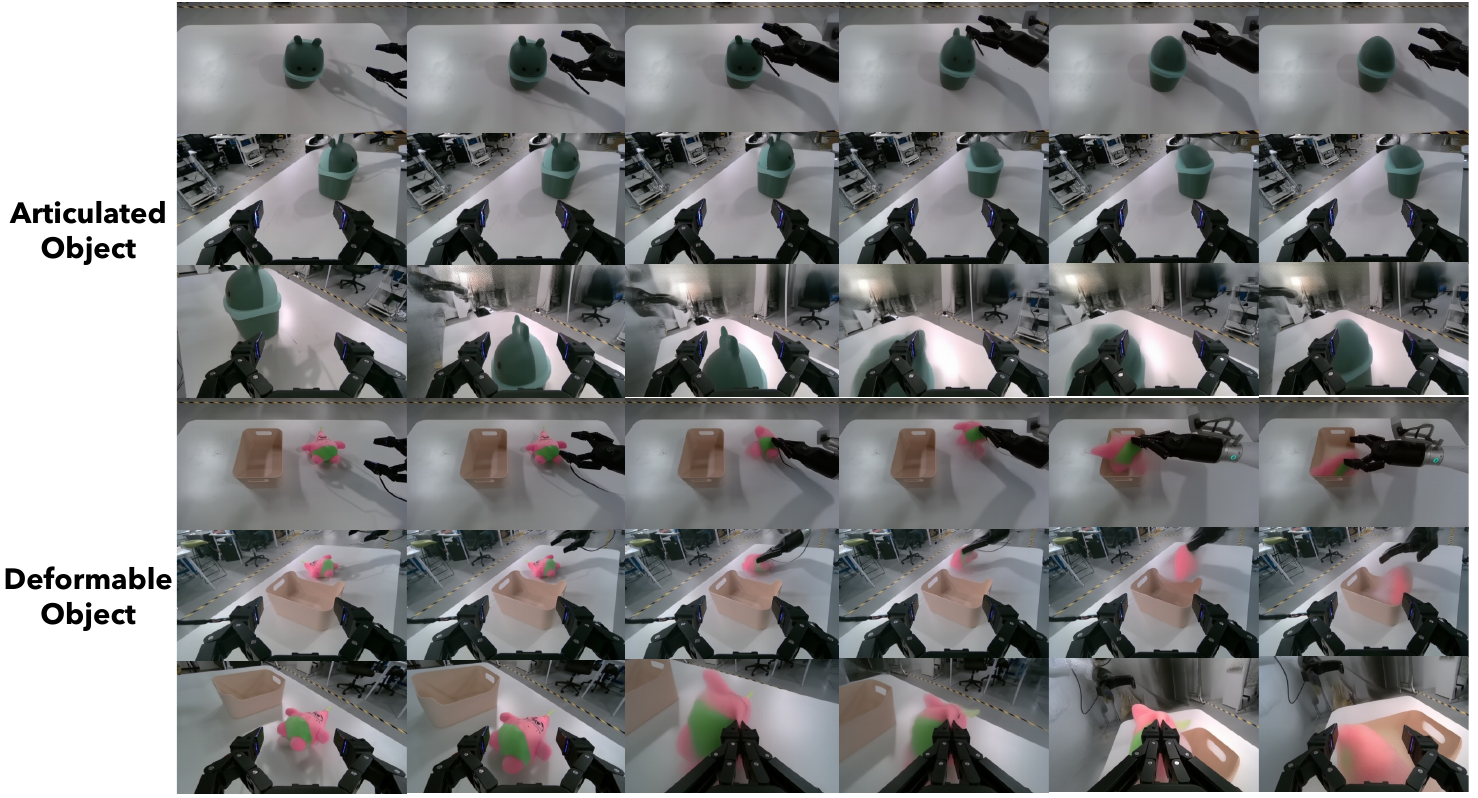}
    \caption{Visualization of Failure Cases.}
    \vspace{-10pt}
    \label{fig:bad_case}
\end{figure}

\section{Additional Visualizations}
\begin{figure*}[t]
    \centering
    \includegraphics[width=\textwidth]{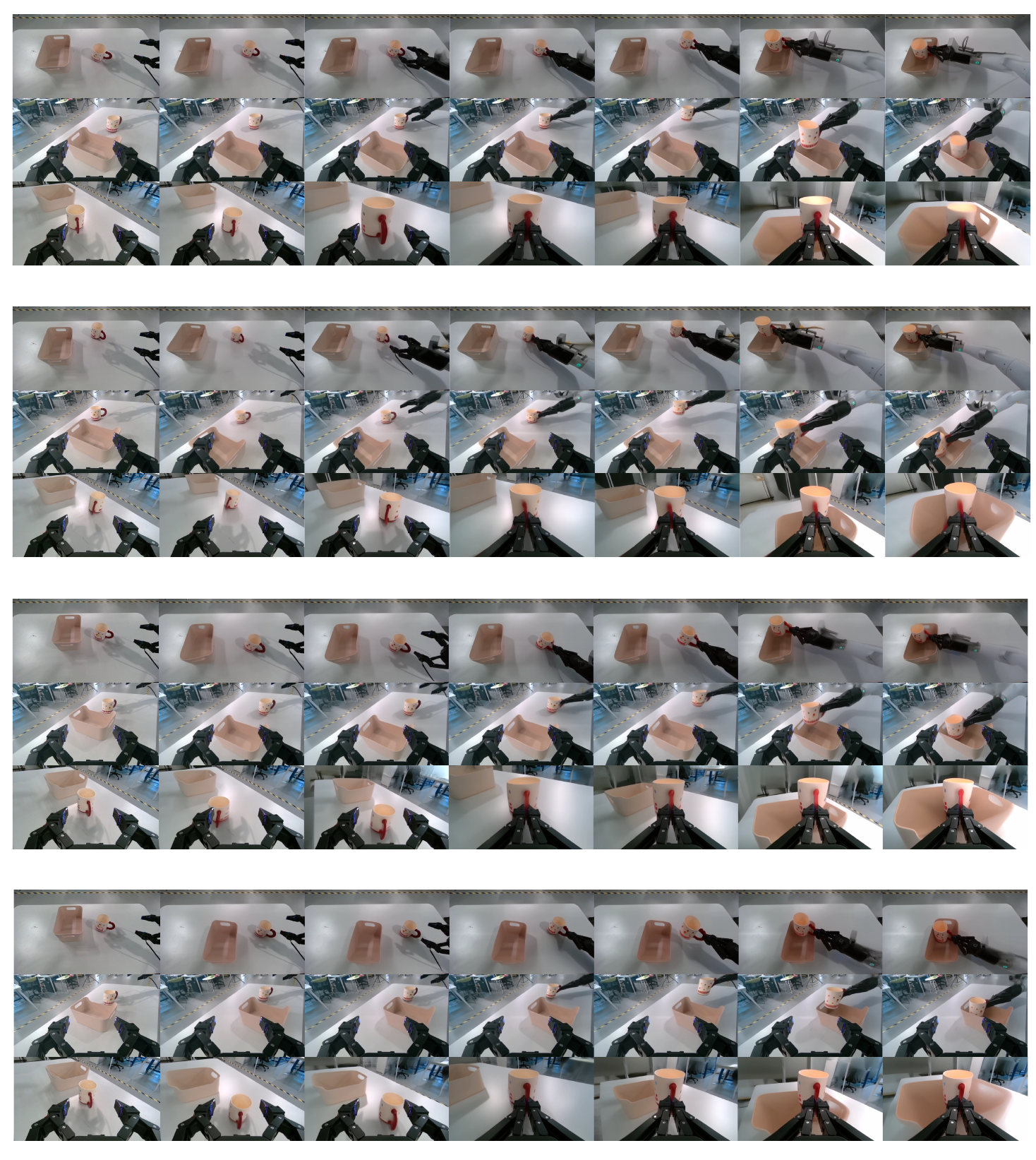}
    \caption{Visualization of Generated Videos on Mug to Basket.}
    \label{fig:supp_vis_mug2box}
\end{figure*}

\begin{figure*}[t]
    \centering
    \includegraphics[width=\textwidth]{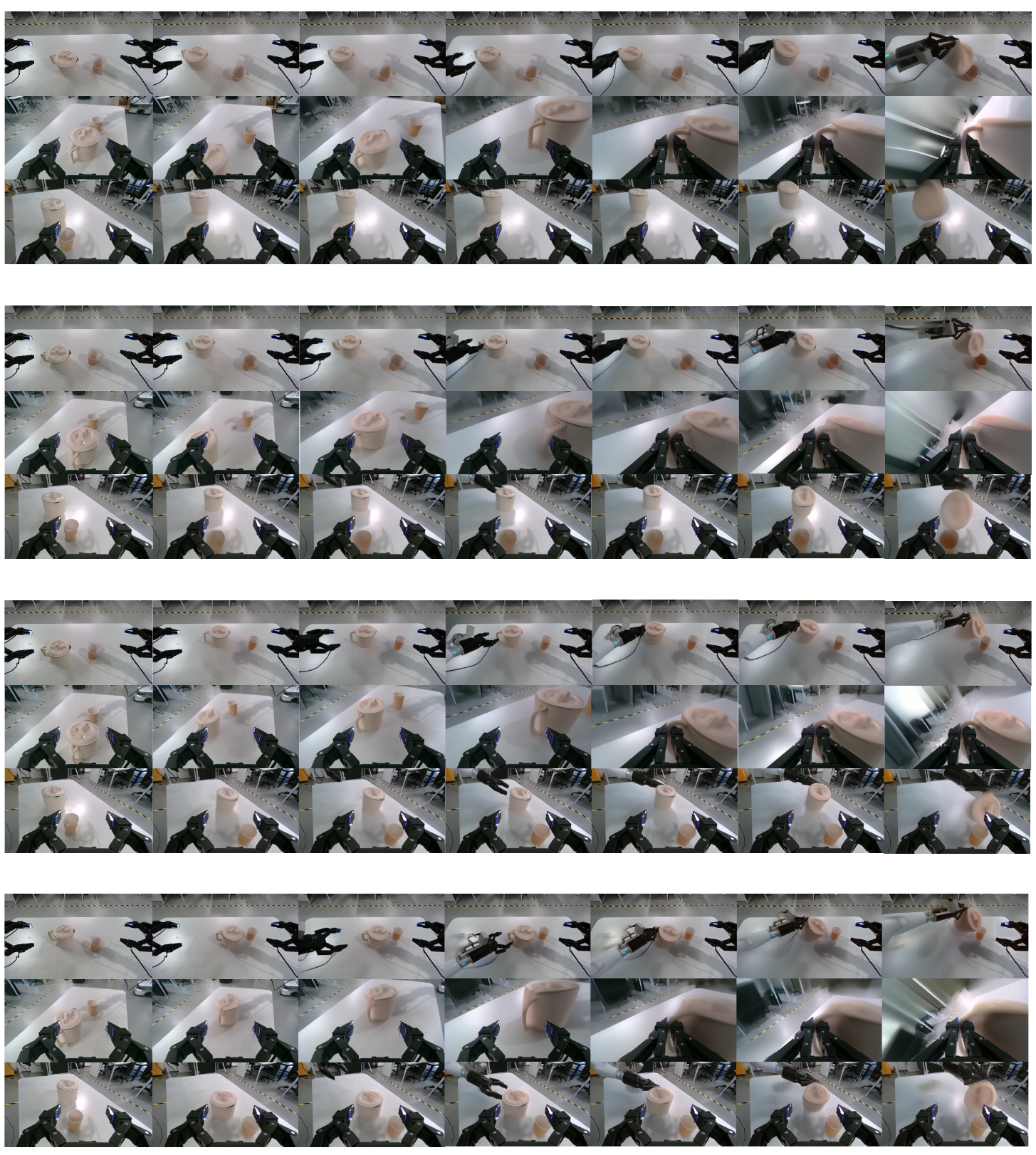}
    \caption{Visualization of Generated Videos on Pour Water.}
    \label{fig:supp_vis_pourwater}
\end{figure*}

\begin{figure*}[t]
    \centering
    \includegraphics[width=\textwidth]{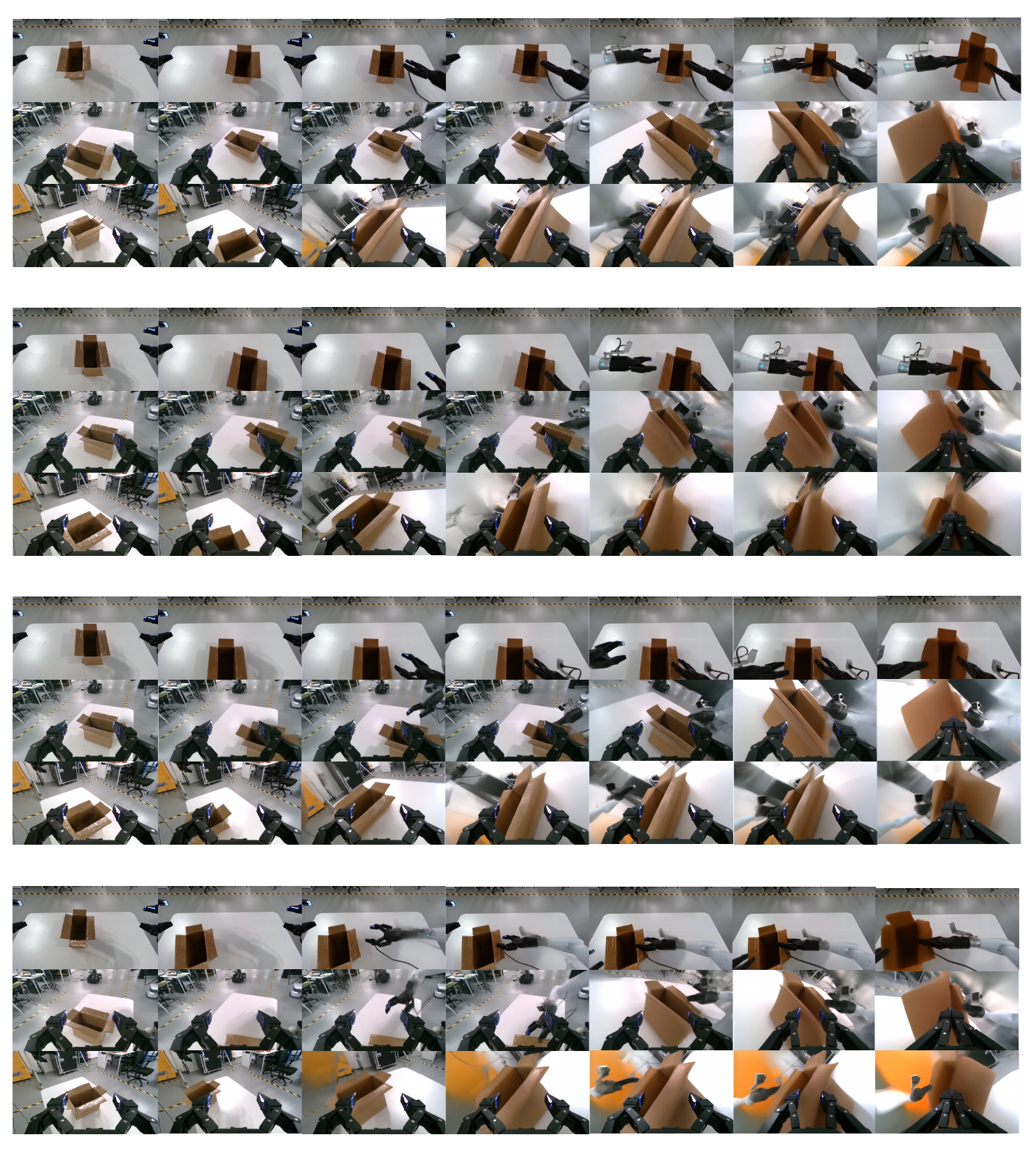}
    \caption{Visualization of Generated Videos on Lift Box.}
    \label{fig:supp_vis_liftbox}
\end{figure*}

\begin{figure*}[t]
    \centering
    \includegraphics[width=\textwidth]{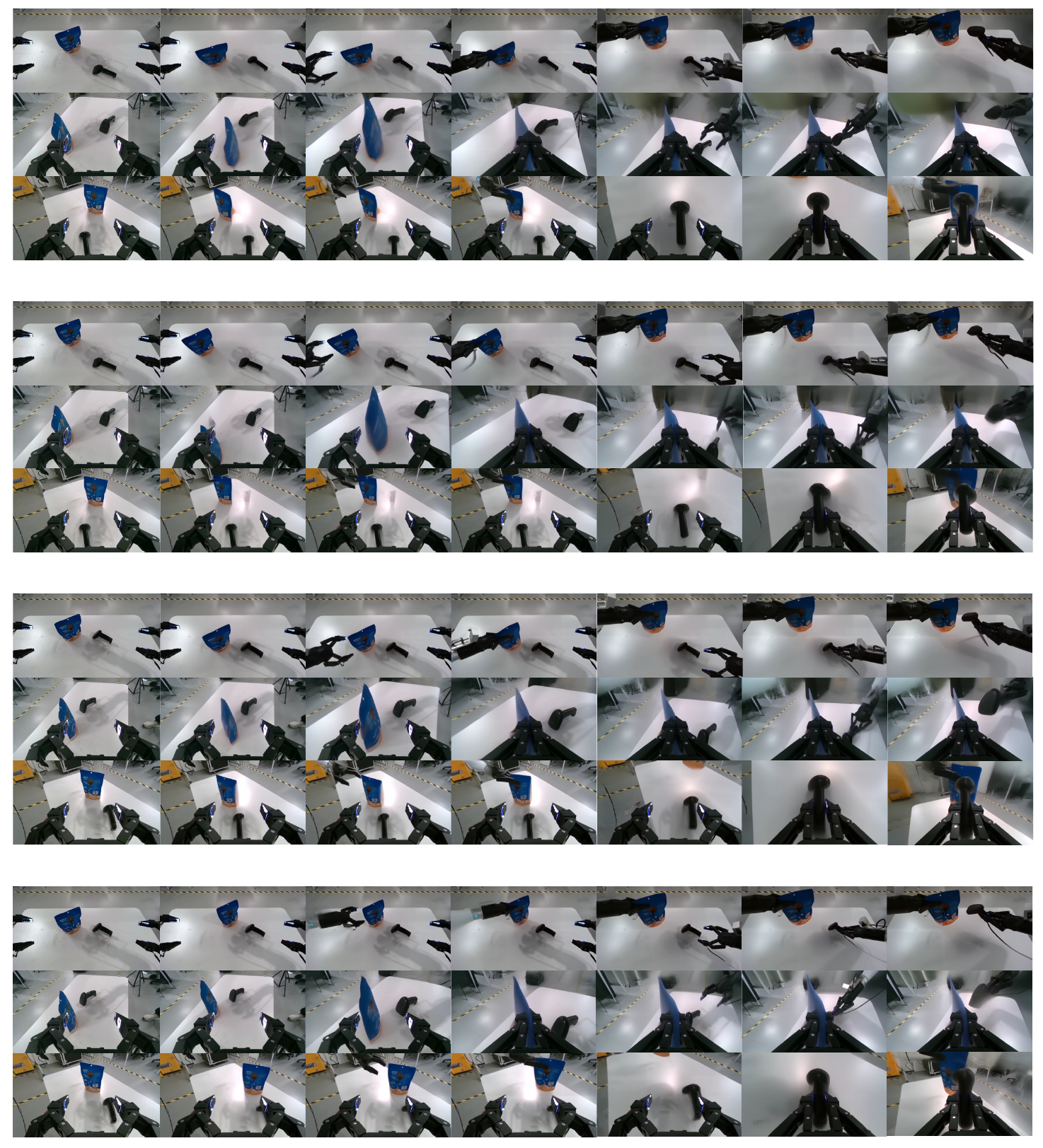}
    \caption{Visualization of Generated Videos on Scan Barcode.}
    \label{fig:supp_vis_scanbarcode}
\end{figure*}

Figure~\ref{fig:supp_vis_mug2box},~\ref{fig:supp_vis_pourwater},~\ref{fig:supp_vis_liftbox},~\ref{fig:supp_vis_scanbarcode} show more visualizations of the four real-world manipulation tasks.

\section{Limitation and Discussion}
Despite the advantages of our proposed Real2Edit2Real framework, which enables scalable multi-view demonstration augmentation, it still has certain limitations. 

\noindent\textbf{Video Generation Time.} As analyzed in Section~\ref{sec:speed}, the video generation module currently poses a computational bottleneck within our framework, particularly in resource-constrained scenarios. Future research could explore the integration of acceleration techniques from the generative modeling community, such as KV caching and model distillation, to further enhance the throughput of our data generation pipeline.

\noindent\textbf{Object Generalization.} As illustrated in Figure~\ref{fig:bad_case}, our generative model exhibits limitations in object generalization, particularly when handling articulated or deformable objects. This stems primarily from the lack of these object categories in our training distribution, which can lead to visual artifacts such as motion blurring or structural inconsistency during video synthesis. To mitigate this, future work could focus on scaling up the diversity and volume of training data to enhance the model's robustness across a broader spectrum of object geometries and physical properties.

\end{document}